\documentclass{article}

\usepackage[main,final]{neurips_2026}

\usepackage[utf8]{inputenc}
\usepackage[T1]{fontenc}
\usepackage[hidelinks]{hyperref}
\usepackage{url}
\usepackage{booktabs}
\usepackage{amsmath,amssymb,amsfonts}
\usepackage{mathtools}
\usepackage{mathrsfs}
\usepackage{amsthm}
\usepackage{graphicx}
\usepackage{subcaption}
\usepackage{float}
\usepackage{microtype}
\usepackage{xcolor}
\usepackage{multirow}
\usepackage{bm}
\usepackage[capitalize,noabbrev]{cleveref}
\DeclareMathOperator{\Expect}{\mathbb{E}}
\DeclareMathOperator{\E}{\mathbb{E}}

\DeclareMathOperator{\Pro}{\mathbb{P}}

\theoremstyle{plain}
\newtheorem{theorem}{Theorem}[section]
\newtheorem{proposition}[theorem]{Proposition}

\newtheorem{lemma}{Lemma}[section]

\newtheorem{property}{Property}[section]
\theoremstyle{definition}

\newtheorem{assumpt}{Assumption}[section]
\theoremstyle{remark}

\newcommand{\clip}{\operatorname{Clip}}

\newcommand{\clipmin}{\operatorname{ClipMin}}

\newcommand{\old}{\text{old}}
\newcommand{\high}{\text{high}}
\newcommand{\low}{\text{low}}

\newcommand{\padtoken}{\mathtt{PAD}}

\title{TIC-GRPO: Provable and Efficient Optimization for Reinforcement Learning from Human Feedback}

\author{
  Lei Pang \\
  Peking University \\
  Beijing, China
  \And
  Jun Luo \\
  The Ohio State University \\
  Columbus, USA
  \And
  Ruinan Jin\thanks{Corresponding author. Most of this work was completed while Ruinan Jin was affiliated with Shugu Yuntai Technology Co., Ltd.} \\
  Mohamed bin Zayed University \\
  of Artificial Intelligence \\
  Abu Dhabi, UAE \\
  \textnormal{jinruinan3@gmail.com}
}

\hypersetup{
  pdftitle={TIC-GRPO: Provable and Efficient Optimization for Reinforcement Learning from Human Feedback},
  pdfauthor={Lei Pang, Jun Luo, Ruinan Jin},
  pdfsubject={Reinforcement learning from human feedback},
  pdfkeywords={TIC-GRPO, policy optimization, importance sampling}
}

\begin{document}

\maketitle

\begin{abstract}
Group Relative Policy Optimization (GRPO) trains language models without a value critic using rewards centered within sampled response groups. We study how importance weighting, clipping, and length normalization affect its stochastic updates and propose Trajectory-level Importance-Corrected GRPO (TIC-GRPO), combining an upper importance-ratio cap with a full-trajectory likelihood ratio. Under common maximum-length normalization, a trajectory change of measure gives a score second moment proportional to the response horizon. We propagate this estimate through the actual empirical baseline and all inner updates. With separately optimized constant steps, the sampling term in TIC's stationarity bound has a linear horizon coefficient, versus order three halves for token-level GRPO$_2$. Matching upper and lower bounds separate the capped methods' worst-case response-level variances by a factor $T$. The bounds track prompt count, response-group size, vocabulary size, reward scale, and score regularity; per-response-normalized GRPO also retains a length-covariance term. Experiments at two Qwen3 scales on four reasoning and coding benchmarks evaluate both mechanisms in sampled-token-normalized training.
\end{abstract}

\section{Introduction}
Reinforcement learning from human feedback (RLHF) aligns language-model outputs with task rewards and human preferences \citep{zhu2023principled,bai2022training}. Standard actor--critic Proximal Policy Optimization (PPO) \citep{schulman2017proximal} uses a value critic; Group Relative Policy Optimization (GRPO) \citep{shao2024deepseekmath} constructs advantages from same-prompt response comparisons, enabling critic-free optimization of reasoning models.

Long responses sampled from an earlier policy make the update's stochastic structure important. Usual GRPO clipping can retain large upward importance ratios on negative-advantage samples. Capping controls this contribution, but clipped token scores generally have nonzero conditional means under the sampling policy. A full-trajectory likelihood ratio changes the entire response law, allowing the raw trajectory-score martingale to control the second moment.

We propose \textsc{TIC-GRPO}, which combines up-only clipping with trajectory-level importance correction and retains empirical group advantages. Its theoretical update uses the maximum response length $T$ as a deterministic denominator. GRPO$_2$ provides a token-level comparator with the same upper cap and denominator. For both methods, the analysis retains the empirical baseline and the dependence of the reward on the entire response.

The trajectory likelihood ratio transfers the TIC score second moment to the current policy, where the trajectory-score martingale gives a factor proportional to $T$. Direct Cauchy--Schwarz across the capped token scores gives a factor proportional to $T^2$ for GRPO$_2$. Our block ascent argument preserves this difference through separate step-size optimization and includes every inner iterate. With fixed reward scale, score regularity, cap, and vocabulary size, the resulting TIC bound is $\mathcal{O}(T/(N-1))+\mathcal{O}(T\sqrt{(B^{-1}+|G|^{-1})/(N-1)})$. The corresponding token-level bound has $T^{3/2}$ in place of $T$ in its square-root term. For original GRPO, normalization by each response's own length introduces an individual-length variance term without an explicit $1/|G|$ factor.

\paragraph{Contributions.}
We establish matching horizon orders for the response-level variance upper and lower bounds of GRPO$_2$ and TIC-GRPO, and an upper-tail instance with growing variance for original GRPO. We connect score moments of the actual empirical-baseline updates to averaged stationarity, accounting for clipping tails and all inner iterates under the specified conditional sampling protocol. We compare TIC-GRPO with DAPO-style GRPO \citep{yu2025dapo} and GSPO \citep{zheng2025group}. GSPO uses a length-normalized sequence importance ratio; TIC-GRPO uses the full trajectory ratio with an upper cap. Section~\ref{sec:gspo-comparison} explains how these differences yield TIC-GRPO's moment-control advantage and compares their empirical update magnitudes. Experiments on Qwen3-1.7B and Qwen3-8B show improved scores on four mathematical-reasoning and coding benchmarks; ablations evaluate both algorithmic components.

\paragraph{Related work.}
Analyses of vanilla policy gradients establish sample-complexity guarantees under moment conditions~\citep{yuan2022general} and study convergence to stationary and locally optimal policies~\citep{zhang2020global}. PPO-Clip theory includes stationary-point convergence with potentially unbounded score functions~\citep{jin2023stationary} and global convergence for deterministic, divergence-regularized algorithms with softmax policies~\citep{liu2025non}. Heavy-tailed policy search relates stationarity and metastability to the policy's tail behavior~\citep{bedi2024sample}. Related adaptive-optimization analyses study Adam~\citep{kingma2014adam}, establishing tight iteration-complexity bounds under smoothness and bounded variance~\citep{wang2023closing} and convergence under moment conditions used for stochastic gradient descent~\citep{DBLP:journals/corr/abs-2410-04458}.

Research on language-model post-training examines alignment faking~\citep{greenblatt2024alignment}, adaptive batch-wise sample scheduling for Direct Preference Optimization~\citep{huang2025adaptive}, and reward models that use evolving policy representations~\citep{huang2026real}. Reasoning efficiency is addressed through sampling that exploits implicit stopping signals~\citep{huang2026does} and hierarchical reinforcement learning that learns sampling temperatures from hidden states~\citep{zhou2026look}. Model efficiency is also studied through retraining-free expert pruning and neuron recombination in sparse mixture-of-experts language models~\citep{zhou2025dropping}. For group-relative methods, related analyses study advantage-estimation bias conditional on groups with mixed rewards~\citep{yang2026your}, while work on diversity preservation in reinforcement learning with verifiable rewards investigates the divergence used for regularization~\citep{li2025choice}.

Applications include dual-clip PPO for complex control in multiplayer online battle arena games~\citep{ye2020mastering}, PPO with self-generated rewards for Chinese spelling correction~\citep{lin2026cec}, and medical visual reasoning optimized with dual-stream GRPO~\citep{zhu2025medeyes}. Heterogeneous agents can share verified rollouts during training and act independently at inference~\citep{zhang2026heterogeneous}. Multimodal modeling also develops dual implicit semantic and spatial memory for vision-language navigation~\citep{zeng2025janusvln}, visual spatio-temporal chains of thought for autonomous driving~\citep{zeng2025futuresightdrive}, and language-guided prototype modulation for diabetic-retinopathy grading~\citep{zhu2025pathology}. Mathematical-reasoning resources include the NuminaMath collection of competition problems and solutions~\citep{li2024numinamath}.

\section{Preliminaries and the GRPO surrogate}
\label{sec:prelim_grpo}

\subsection{Reinforcement learning setup}
\label{sec:rl_setup}

The policy parameter ranges over $\theta\in\mathbb R^l$. The prompt set is $C$; the prompt law $p$ and terminal reward function are independent of this parameter. Responses use a finite vocabulary $\mathcal V$ and a positive integer horizon $T$. Let \(c\) be a prompt and let \(a_t\in\mathcal V\) denote the
\(t\)-th non-padding response token before termination. We write
\(s_t=(c,a_{\le t})\) for the prefix state conditioned on the prompt.
For notational convenience, variable-length responses are embedded in
a fixed-horizon representation by appending a distinguished padding
symbol \(\padtoken\), which is masked out in all token-level sums:
\[
s_T=(c,a_1,\ldots,a_t,\underbrace{\padtoken,\ldots,\padtoken}_{T-t\ \mathrm{positions}}),\qquad |s_T|=t.
\]
Here \(\mathcal S_T\) is the set of padded terminal responses and \(|s_T|\) counts active response tokens. Padding after termination is deterministic, not a random action: its transition probability is one and its score is zero. Sums up to \(T\) are therefore masked over active tokens. Active transitions satisfy \(\Pro_\theta(s_t\mid s_{t-1})=\pi_\theta(a_t\mid s_{t-1})\), and terminal response probabilities factorize as
\[
\Pro_\theta(s_T\mid c)
=
\prod_{t=1}^{|s_T|}\Pro_\theta(s_t\mid s_{t-1})
=
\prod_{t=1}^{T}\Pro_\theta(s_t\mid s_{t-1}),
\]
where the second product includes only unit-probability padding factors
after termination. Rewards satisfy $|r(s_T)|\leq R$. The objective and policy gradient are
\begin{align}
J(\theta)
&:=
\Expect_{c\sim p,\,s_T\sim\Pro_\theta(\cdot\mid c)}[r(s_T)],
\qquad J^*:=\sup_\theta J(\theta)<\infty,
\label{Value_Function}\\
\nabla J(\theta)
&=
\Expect_{c\sim p,\,s_T\sim\Pro_\theta(\cdot\mid c)}
\!\left[\nabla\log\Pro_\theta(s_T\mid c)\,r(s_T)\right].
\label{gradient}
\end{align}

Throughout, \(\nabla\) denotes gradients with respect to \(\theta\), unless stated otherwise.

\subsection{Review of group relative policy optimization}
\label{sec:grpo_review}

This subsection fixes the GRPO-style token-level surrogate whose
second moment will be analyzed later. Under a fixed old policy
\(\pi_{\theta_{\old}}\), each inner batch contains $B$ prompt occurrences drawn independently from $p$, conditional on the preceding iterates. Each occurrence of prompt $c$ is assigned a group $G_c$ of $|G|$ responses that are conditionally independent given the sampled prompts and the preceding iterates, with law $\Pro_{\theta_{\old}}(\cdot\mid c)$. These prompt and response samples are fresh at each inner update. Let \(\xi_c\) be the empirical measure
on \(\mathcal S_T\): if \(s_T\) appears \(w\) times in \(G_c\), then
\begin{equation}
\label{eq:group_empirical_measure}
\xi_c(s_T)=w/|G|,
\end{equation}
where the group size is common across prompts. The group-relative
advantage is
\[
A_c(s_T)=r(s_T)-\mu_c,
\qquad
\mu_c=\sum_{s_T\in\mathcal S_T}\xi_c(s_T)r(s_T).
\]
Following recent GRPO-style implementations~\citep{liu2025understanding,chu2025gpg,dai2025s,liu2025deepseek}, we omit standard-deviation normalization of group-relative advantages: it can introduce additional bias \citep{liu2025understanding,chu2025gpg} and is unnecessary for our variance comparison.

The response batch size is the total number of sampled responses per inner mini-batch: $B|G|$, from $B$ prompts with $|G|$ responses each. For prompt-only functions, the empirical prompt measure $\zeta_k(c)$ is the prompt's occurrence count divided by $B$. Each occurrence, including repeated prompts, retains its own $G_c$, $\xi_c$, and $\mu_c$. In occurrence sums, $c$ denotes that occurrence's prompt; its group quantities are evaluated before averaging the $B$ contributions. GRPO maximizes the group-relative surrogate
\begin{align}
\label{eq:grpo_clipmin}
\clipmin(s_t,\theta,\theta_{\old})
:= \min\Bigg\{&
\frac{\Pro_\theta(s_t\mid s_{t-1})}
     {\Pro_{\theta_{\old}}(s_t\mid s_{t-1})}
A_c(s_T),\notag\\
&
\clip\!\left(
\frac{\Pro_\theta(s_t\mid s_{t-1})}
     {\Pro_{\theta_{\old}}(s_t\mid s_{t-1})},
\epsilon_\low,\epsilon_\high
\right)A_c(s_T)
\Bigg\},
\end{align}
where
\(\clip(x,\epsilon_\low,\epsilon_\high)
:=\min\{\max\{x,1-\epsilon_\low\},1+\epsilon_\high\}\).
The corresponding inner objective is
\begin{align}
\label{eq:grpo_objective}
\mathcal{L}^{(k)}_{\mathrm{GRPO}}(\theta,\theta_{\old})
:=
\frac1B\sum_{\text{prompt occurrences}}
\sum_{s_T\in\mathcal S_T}\xi_c(s_T)
\frac{1}{|s_T|}
\sum_{t=1}^{T}
\clipmin(s_t,\theta,\theta_{\old}).
\end{align}
The score-form GRPO direction is $\widehat G_{\mathrm{GRPO}}^{(k)}=\nabla\mathcal L_{\mathrm{GRPO}}^{(k)}$, with any subgradient between the one-sided values at clipping thresholds. The sampled rewards and empirical advantages are held fixed in this differentiation.
The old policy is held fixed during the inner loop and refreshed after
\(K\) updates. The sampled termination action retains its score when its probability depends on the parameter. Only the subsequent deterministic padding coordinates have zero score, so the sum to $T$ is equivalent to summing over active tokens.

DAPO uses asymmetric clipping thresholds and omits the reference-policy KL penalty \citep{yu2025dapo}. These choices are orthogonal to the variance mechanism: the surrogate remains token-level, with no uniform upper-tail cap on the negative-advantage branch. When $A_c(s_T)<0$, Eq.~\eqref{eq:grpo_clipmin} can select the unclipped ratio, admitting large upward ratios into the gradient. This incomplete upper-tail control is the first variance bottleneck addressed below. We use ``GRPO'' for the critic-free token-level mechanism shared by GRPO and DAPO in the theory.

\subsection{Stochastic update structure}
\label{sec:what_control}
The empirical baseline, clipping multiplier, and response length are functions of the same sampled data. Their dependence is retained in the update analysis. TIC uses the trajectory-score martingale under its own policy law, together with a full-trajectory change of measure. GRPO$_2$ uses single-token conditional moments and Cauchy--Schwarz across the token positions.

\section{Trajectory-level importance-corrected GRPO (TIC-GRPO)}
\label{sec:tic_grpo}

GRPO combines token-level importance weighting with clipping. TIC-GRPO first caps upper-tail weights by up-only clipping, then replaces token-level weights by the full trajectory likelihood ratio. The intermediate comparator GRPO$_2$ retains token-level weighting with the same up-only cap and maximum-length denominator as TIC-GRPO.

\paragraph{Up-only clipping.}
For a token \(s_t\) on a trajectory \(s_T\), define the one-step ratio \(\rho_t(\theta,\theta_{\old}) := \frac{\Pro_{\theta}(s_t\mid s_{t-1})} {\Pro_{\theta_{\old}}(s_t\mid s_{t-1})}.\) The standard GRPO clipping rule does not uniformly cap upper-tail
ratios. Indeed, when \(A_c(s_T)<0\) and
\(\rho_t(\theta,\theta_{\old})>1+\epsilon_{\high}\), the minimum in
the GRPO surrogate can select the unclipped term
\(\rho_t(\theta,\theta_{\old})A_c(s_T)\). We therefore use the
up-only clipped ratio \(\bar\rho_t(\theta,\theta_{\old}) := \min\{\rho_t(\theta,\theta_{\old}),1+\epsilon_{\high}\}.\) This clipping rule caps large upward ratios regardless of the sign of
the advantage.

\paragraph{Intermediate comparator: GRPO$_2$.}
GRPO$_2$ uses up-only clipped token ratios and maximum-length normalization. Its \(k\)-th inner score-form update is
\begin{align}
\label{eq:grpo2_update_direction}
\widehat G_{\mathrm{GRPO}_2}^{(k)}(\theta,\theta_{\old})
:=
\frac{1}{T}
\frac1B\sum_{\text{prompt occurrences}}
\sum_{s_T\in\mathcal S_T}\xi_c(s_T)
\sum_{t=1}^{T}
\bar\rho_t(\theta,\theta_{\old})A_c(s_T)
\nabla\log\Pro_{\theta}(s_t\mid s_{t-1}),
\end{align}
where token sums are masked after termination and $T$ is the maximum response length. The common cap and denominator isolate the effect of trajectory-level importance correction when comparing GRPO$_2$ with TIC-GRPO.

\paragraph{Token-score aggregation.}
The GRPO$_2$ feature sums capped token scores. Changing the next-token distribution bounds each token's conditional second moment under its old-policy prefix law. Cauchy--Schwarz across the $T$ positions controls the complete feature, including all cross terms, yielding the moment estimate used in the convergence comparison.

\paragraph{Trajectory-level importance correction.}
TIC-GRPO replaces the token-level ratios by the trajectory likelihood
ratio \(\rho_{0:T}(\theta,\theta_{\old}) := \frac{\Pro_{\theta}(s_T\mid c)} {\Pro_{\theta_{\old}}(s_T\mid c)}.\) The untruncated ratio \(\rho_{0:T}\) is the Radon--Nikodym derivative
for changing measure from old-policy trajectories to current-policy
trajectories, and it pairs with the full trajectory score
\[
G_T^\theta
:=
\nabla\log\Pro_{\theta}(s_T\mid c)
=
\sum_{t=1}^{T}
\nabla\log\Pro_{\theta}(s_t\mid s_{t-1}).
\]
TIC-GRPO applies the same up-only clipping principle at the trajectory
level: \(\bar\rho_{0:T}(\theta,\theta_{\old}) := \min\{\rho_{0:T}(\theta,\theta_{\old}),1+\epsilon_{\high}\}.\) The clipped ratio is no longer an exact Radon--Nikodym derivative, but retains the needed likelihood structure: one copy of \(\rho_{0:T}\) changes measure while the cap controls the remaining weight. The second moment thus reduces to a current-policy trajectory-score second moment, where martingale cancellation applies.

\paragraph{TIC-GRPO update.}
The \(k\)-th inner TIC-GRPO score-form direction is
\begin{align}
\label{eq:tic_update_direction}
\widehat G_{\mathrm{TIC}}^{(k)}(\theta,\theta_{\old})
:=
\frac{1}{T}
\frac1B\sum_{\text{prompt occurrences}}
\sum_{s_T\in\mathcal S_T}\xi_c(s_T)
\bar\rho_{0:T}(\theta,\theta_{\old})A_c(s_T)
\nabla\log\Pro_{\theta}(s_T\mid c).
\end{align}
The parameter update is \(\theta_{k+1} = \theta_k+\eta\,\widehat G_{\mathrm{TIC}}^{(k)} (\theta_k,\theta_{\old}),\) with the old policy held fixed during the inner loop and refreshed
after \(K\) updates, as in GRPO. Adam is widely used in practice~\citep{kingma2014adam,jin2026generalized}; our analysis uses stochastic gradient ascent to isolate the RL update mechanisms.

Both capped updates use the maximum-length denominator $T$; responses may terminate earlier, and only active scores enter the numerator. Defined directly by~\eqref{eq:grpo2_update_direction} and~\eqref{eq:tic_update_direction}, both score-form directions retain the capped multiplier at every ratio, including above the cap.

\section{Variance analysis and martingale structure}
\label{sec:variance_analysis}
We bound response-level variance cores above and below, then combine batch variances with mean-update stability. The upper bounds isolate clipping and trajectory-level importance weighting; binary constructions determine attainable horizon dependence.

Fix an arbitrary initial parameter $\theta_{1,0}\in\mathbb R^l$. We use the following outer--inner notation. The old policy is refreshed
at the beginning of each subsequent outer iteration, \(\theta_{n,0}=\theta_{n-1,K},\qquad n\geq2,\) and \(k=0,\ldots,K-1\) indexes the inner updates performed under the
fixed old policy \(\theta_{n,0}\). For an algorithm \(\mathsf A\in\{\mathrm{GRPO},\mathrm{GRPO}_2,\mathrm{TIC}\},\) we write the update in the common score-form notation \(\theta_{n,k+1} = \theta_{n,k} + \eta\,\widehat G_{\mathsf A}^{(n,k)}.\) Here \(\widehat G_{\mathsf A}^{(n,k)}\) is the corresponding stochastic update direction: original GRPO uses its clipping-threshold subgradient, and the two capped methods use the score-form multipliers in Section~\ref{sec:tic_grpo}. All mean and moment estimates retain the empirical baseline. The outer and within-block information fields are
\[
\mathscr F_n=\sigma(\theta_{1,0},\ldots,\theta_{n,0}),
\qquad
\mathscr F_{n,k}
=
\sigma(\mathscr F_n,\theta_{n,1},\ldots,\theta_{n,k}).
\]

We use the following assumptions throughout the variance and convergence
analysis.

\begin{assumpt}[\textbf{Lipschitz score in \(\theta\)}]
\label{L_smooth}
There exists \(L>0\) such that for every active token transition
\(s_{t-1}\to s_t\) and all $\theta,\theta'\in\mathbb R^l$,
\[
\big\|
\nabla \log \Pro_{\theta}(s_t\mid s_{t-1})
-
\nabla \log \Pro_{\theta'}(s_t\mid s_{t-1})
\big\|
\le
L\|\theta-\theta'\|.
\]
The deterministic padding transitions have zero score and satisfy this
condition trivially.
\end{assumpt}

\begin{assumpt}[\textbf{Bounded rewards and common support}]
\label{bounded_reward}
There exists \(R>0\) such that \(|r(s_T)|\le R\) for all terminal
states \(s_T\). Moreover, all likelihood ratios and logarithms appearing
in the analysis are evaluated on a common positive support: for every
active transition on this support and every $\theta\in\mathbb R^l$,
\[
\Pro_\theta(s_t\mid s_{t-1})>0,
\qquad
\Pro_{\theta_{n,0}}(s_t\mid s_{t-1})>0 .
\]
Transitions after termination are deterministic and have zero score; a sampled termination action is included in the likelihood. All response lengths satisfy
\(1\le |s_T|\le T\).
\end{assumpt}

Related policy-gradient analyses impose regularity on log policies and their scores~\citep{zhang2023model,xiong2021non,ding2021global,suk2025optimization}. Here the score itself need not be uniformly bounded. Common support ensures that the ratios and logarithms are well defined.

The theoretical sampling protocol is specified as follows.
Conditional on $\mathscr F_{n,k}$, the $B\geq1$ prompt occurrences in an inner batch are independent draws from $p$, each contributing weight $1/B$. Thus the empirical prompt measure satisfies \(\Expect[\zeta_k(c)\mid\mathscr F_{n,k}]=p(c).\) Given the sampled prompts and $\mathscr F_{n,k}$, the $|G|$ responses within each occurrence are independent draws from $\Pro_{\theta_{n,0}}(\cdot\mid c)$. The response groups are freshly sampled at each inner update. Dependence between different response groups is allowed. Each empirical baseline and sampled response length belong to the same group.
This conditional sampling protocol is used throughout the stationarity analysis. We use $\Expect_\theta$ for the joint expectation over $c\sim p$ and $s_T\sim\Pro_\theta(\cdot\mid c)$; a displayed conditioning on $c$ retains only the response expectation. The prompt reward mean is \(\mu_{\theta,c}=\Expect_\theta[r(s_T)\mid c].\) The cap satisfies $\epsilon_{\high}\geq0$; the GRPO drift bound with reciprocal clipping margins is stated for positive margins.

\subsection{Trajectory scores and martingale cancellation}
\label{sec:martingale_structure}

Fix a prompt \(c\) and a policy parameter \(\theta\). Define the
one-step score and the full trajectory score as
\[
g_t^\theta
:=
\nabla\log\Pro_\theta(s_t\mid s_{t-1}),
\qquad
G_T^\theta
:=
\nabla\log\Pro_\theta(s_T\mid c)
=
\sum_{t=1}^{T}g_t^\theta,
\]
where the sum is masked after termination. Under the current-policy
trajectory law \(\Pro_\theta(\cdot\mid c)\), the one-step score terms
form a martingale-difference sequence: \(\E_\theta[g_t^\theta\mid s_{t-1}]=0.\) Consequently, the cross terms vanish in the squared trajectory score.

\begin{property}[Trajectory-score second moment]
\label{prop:martingale_identity}
For any prompt \(c\) and any \(\theta\),
\[
\Expect_{s_T\sim\Pro_\theta(\cdot\mid c)}
\!\left[
\|G_T^\theta\|^2
\right]
=
\Expect_{s_T\sim\Pro_\theta(\cdot\mid c)}
\!\left[
\sum_{t=1}^{T}\|g_t^\theta\|^2
\right].
\]
\end{property}

For TIC, the full-trajectory likelihood ratio changes the sampling law before this identity is applied. For GRPO$_2$, the update core is bounded by Cauchy--Schwarz across token positions and a separate next-token change of measure at each prefix.

\subsection{Variance cores for GRPO, \texorpdfstring{GRPO$_2$}{GRPO2}, and TIC-GRPO}
\label{sec:variance_cores}

Let
\[
\rho_t
:=
\frac{\Pro_{\theta_{n,k}}(s_t\mid s_{t-1})}
     {\Pro_{\theta_{n,0}}(s_t\mid s_{t-1})},
\qquad
\rho_{0:T}
:=
\frac{\Pro_{\theta_{n,k}}(s_T\mid c)}
     {\Pro_{\theta_{n,0}}(s_T\mid c)},
\qquad
c_+:=1+\epsilon_{\high}.
\]
We write the up-only clipped token and trajectory ratios as \(\rho_t^+ := \min\{\rho_t,c_+\}, \qquad \rho_{0:T}^+ := \min\{\rho_{0:T},c_+\}.\) For original GRPO, let \(m_t^{\mathrm{GRPO}}\) denote the multiplier
that appears in the score-form core of the original clipped token-level
update. Away from the clipping thresholds, this multiplier is either
zero or the active token ratio selected by the GRPO clipping rule; at a
threshold, we take any subgradient in the closed interval between the
two one-sided limits. Because the negative-advantage branch can select
an unclipped upper-tail ratio, the multiplier is controlled on the
sampled old-policy support only through
\begin{equation}
\label{eq:upper_tail_multiplier}
|m_t^{\mathrm{GRPO}}|\le \mathcal M_n,
\qquad
\mathcal M_n
:=
\max_{s_T\in\mathcal S_T^{(n)}}
\max_{1\le t\le |s_T|}
\frac{1}{\Pro_{\theta_{n,0}}(s_t\mid s_{t-1})}.
\end{equation}
Here $\mathcal S_T^{(n)}$ contains the responses sampled during outer iteration $n$, including their prompt labels. The factor $\mathcal M_n$ is a random maximum over their active transitions and controls the upper-tail contribution in GRPO.

We compare the following stochastic update cores:
\begin{equation}
\label{eq:grpo_core}
\begin{gathered}
Z_{\mathrm{GRPO}}\coloneqq\frac{A_c(s_T)}{|s_T|}\sum_{t=1}^{|s_T|}m_t^{\mathrm{GRPO}}g_t^{\theta_{n,k}},\qquad
Z_{\mathrm{GRPO}_2}\coloneqq\frac{A_c(s_T)}{T}\sum_{t=1}^{T}\rho_t^+g_t^{\theta_{n,k}},\\
Z_{\mathrm{TIC}}\coloneqq\frac{A_c(s_T)}{T}\rho_{0:T}^+G_T^{\theta_{n,k}}.
\end{gathered}
\end{equation}
Each core is a single response's contribution to the batch update direction. Empirical averaging over prompts and responses gives \(\widehat G_{\mathsf A}^{(n,k)}\). The direct score-form update therefore satisfies \(\|\theta_{n,k+1}-\theta_{n,k}\|_2^2=\eta^2\|\widehat G_{\mathsf A}^{(n,k)}\|_2^2\), the squared Euclidean distance between consecutive parameter iterates. By Jensen's inequality, each conditional core second-moment upper bound, multiplied by \(\eta^2\), also bounds the corresponding conditional expected squared distance.

These three cores isolate the variance mechanisms of interest:
original GRPO may retain uncontrolled upper-tail multipliers, GRPO$_2$
caps upper-tail ratios but remains token-level, and TIC-GRPO applies
the same upper-tail cap at the trajectory level.

\subsection{Upper bounds for the three variance cores}
\label{sec:variance_upper_bounds}
\begin{proposition}[Core update moments, informal]
\label{prop:variance-upper-informal}
Under Assumptions~\ref{L_smooth}--\ref{bounded_reward} and the stated sampling protocol, the conditional second moments of $Z_{\mathrm{GRPO}}$, $Z_{\mathrm{GRPO}_2}$, and $Z_{\mathrm{TIC}}$, given $\mathscr F_{n,k}$, are respectively
\[
 \mathcal{O}\!\left(R^2L\Expect[\mathcal M_n^2\mid\mathscr F_{n,k}]\right),
 \qquad \mathcal{O}(R^2Lc_+\log|\mathcal V|),
 \qquad \mathcal{O}\!\left(\frac{R^2Lc_+\log|\mathcal V|}{T}\right).
\]
The constants are universal; the first bound is interpreted in the extended real sense. The distinguished response uses the empirical prompt and response weights, and the same bounds apply to the empirical average update by Jensen's inequality.
\end{proposition}
Proposition~\ref{prop:variance_upper_bounds} in Appendix~\ref{app:proof_variance_upper_bounds} gives the full statement with explicit constants and its proof.

Each displayed second-moment bound also bounds the corresponding conditional variance, since subtracting the conditional mean reduces the second moment. With $R,L,c_+$ and $|\mathcal V|$ fixed, the capped response-level cores therefore have variance upper bounds of order $1$ for GRPO$_2$ and $1/T$ for TIC-GRPO. For original GRPO, the dependence on the old-policy inverse probabilities remains explicit through $\mathcal M_n$.

For TIC-GRPO the key inequality is $(\rho_{0:T}^+)^2\leq c_+\rho_{0:T}$. The trajectory ratio changes the law to the current policy, under which Property~\ref{prop:martingale_identity} applies. These bounds describe the two importance-weighting mechanisms under a common denominator. The convergence analysis additionally separates the prompt and response contributions to the batch variance and controls the complete mean update within each outer block. 

\subsection{Lower bounds and tightness of the variance scaling}
\label{sec:variance_lower_bounds}

The horizon dependence of the capped response-level bounds is attained by binary policies with bounded rewards. The constructions retain the empirical group baseline and the normalizations in the three expressions of Eq.~\eqref{eq:grpo_core}.

\begin{proposition}[Response-level variance lower bounds, informal]
\label{prop:variance-lower-informal}
For every $R,L>0$, $c_+\geq1$, $T\geq2$, and $|G|\geq2$, separate one-dimensional binary policy instances satisfying Assumptions~\ref{L_smooth}--\ref{bounded_reward}, with one prompt, independent responses, and deterministic length $T$, attain conditional response-core variances
\[
 \mathrm{GRPO}_2:\quad\mathcal{\mathrm{\Omega}}\!\left(R^2L\left(1-\frac1{|G|}\right)\right),
 \qquad
 \mathrm{TIC}:\quad\mathcal{\mathrm{\Omega}}\!\left(\frac{R^2L}{T}\left(1-\frac1{|G|}\right)\right).
\]
Original GRPO has a binary instance with variance $\mathcal{\mathrm{\Omega}}(R^2LT)$ when $|G|=2$, $T\geq64$, and $T^2>2c_+$. The constants are universal. All cores are evaluated at fixed old/current parameter pairs before sampling. The old-policy success probabilities are $T^{-2}$ for original GRPO and $1/(4c_+)$ for GRPO$_2$.
\end{proposition}
Proposition~\ref{prop:hard_instance_variance} in Appendix~\ref{app:hard_instance_variance} gives the explicit lower bounds and their complete constructions.

Together, Propositions~\ref{prop:variance-upper-informal} and~\ref{prop:variance-lower-informal} give matching horizon orders $1$ and $1/T$ for the worst-case response-level variances of GRPO$_2$ and TIC-GRPO, respectively, with fixed $|G|\geq2$ and a binary vocabulary. Equivalently, the second moments of their unnormalized numerators have orders $T^2$ and $T$. Original GRPO admits a variance of order at least $T$ as its old-policy upper-tail ratios grow, and the same construction gives an unnormalized second moment of order at least $T^3$. These statements concern the individual response cores; the empirical batch averages and the iterates generated by the algorithms are analyzed in Section~\ref{sec:convergence_consequences}.

\subsection{Empirical update magnitudes}
\label{sec:variance_diagnostics}

To examine update magnitudes in practical LLM fine-tuning, we track
\(\|\theta_{k+1}-\theta_k\|_2^2\), the squared Euclidean distance between consecutive parameter iterates, on Qwen3-0.6B and Qwen3-1.7B. For a fixed learning rate and a direct stochastic-gradient update, its expectation equals the batch update direction's second moment multiplied by the squared learning rate. In the experiments, this distance is computed from the actual parameter iterates produced by the optimizer. All compared methods use the same data and
optimizer settings from Appendix~\ref{sec:experiments_setup}.

\begin{figure}[t]
    \centering
    \begin{subfigure}[t]{0.47\linewidth}
        \centering
        \vspace{0pt}
        \includegraphics[width=\linewidth,trim=7bp 7bp 7bp 7bp,clip]{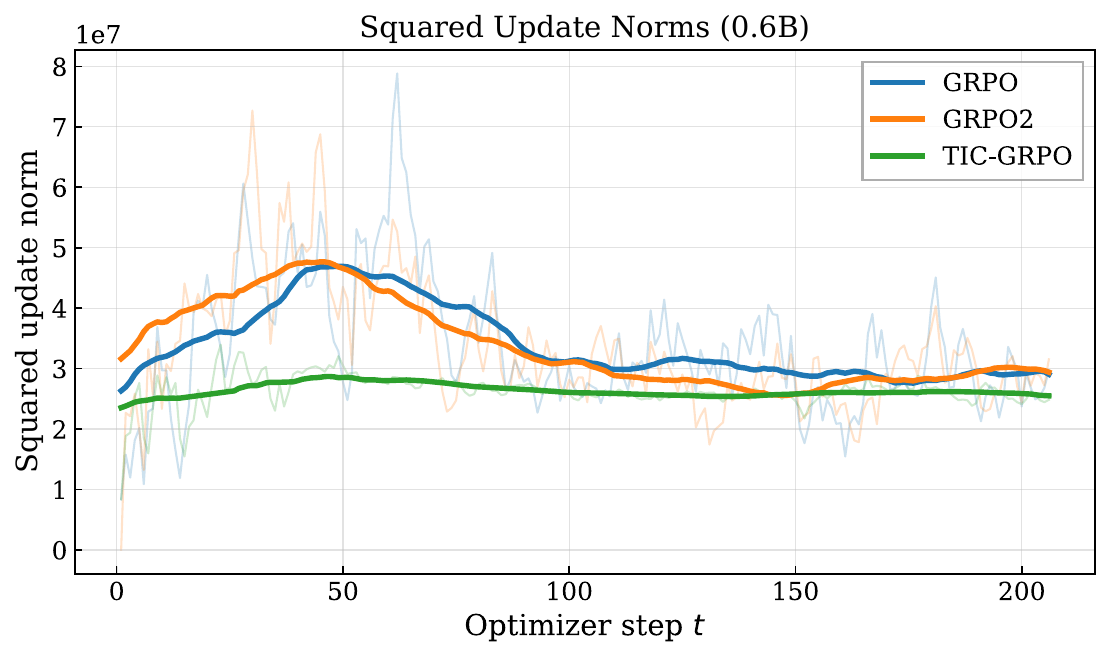}
        \caption{Qwen3-0.6B.}
    \end{subfigure}
    \hfill
    \begin{subfigure}[t]{0.47\linewidth}
        \centering
        \vspace{0pt}
        \includegraphics[width=\linewidth,trim=7bp 7bp 7bp 7bp,clip]{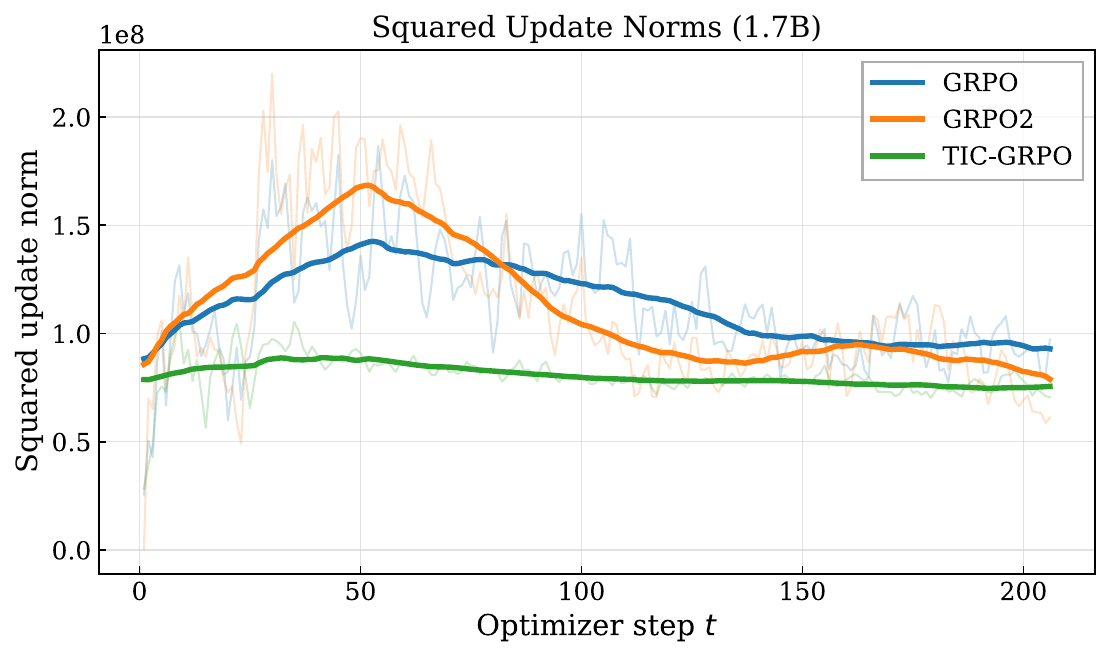}
        \caption{Qwen3-1.7B.}
    \end{subfigure}
    \caption{Squared Euclidean distances between consecutive parameter iterates, $\|\theta_{k+1}-\theta_k\|_2^2$. TIC-GRPO has the lowest and most stable values in these runs.}
    \label{fig:variance_diagnostics}
\end{figure}

The curves measure realized update sizes under the experimental optimizer and provide an empirical comparison of the three training rules.

\subsection{Comparison with GSPO}
\label{sec:gspo-comparison}

GSPO applies PPO-style minimum clipping to the root-normalized sequence ratio $\rho_{0:T}^{1/|s_T|}$. Negative advantages can retain arbitrarily large root ratios above the upper clipping threshold. TIC-GRPO instead satisfies $(\rho_{0:T}^+)^2\leq c_+\rho_{0:T}$ regardless of the advantage's sign: the cap controls one weight factor, and the exact trajectory ratio transfers the remaining score moment to the current policy, where Property~\ref{prop:martingale_identity} applies. GSPO's root ratio generally differs from this density ratio. Even if the root ratio were capped, applying the same rescaling-and-transfer argument would leave $\rho_{0:T}^{1/|s_T|-1}$ inside the current-policy expectation, which is unbounded as $\rho_{0:T}\to0$ for $|s_T|>1$. TIC combines the two properties supporting the uniform moment reduction in Proposition~\ref{prop:variance-upper-informal}.

Matched runs with the same data, optimizer, and training settings over steps 1--200 on Qwen3-0.6B and Qwen3-1.7B give TIC-GRPO about 21\% and 26\% lower temporal means of the realized squared update norm than GSPO, respectively. The corresponding standard deviations across training steps are about 69\% and 76\% lower, indicating smaller and more stable realized updates. Appendix~\ref{sec:gspo-experiments} reports the protocol and interval-level results in Table~\ref{tab:gspo-update-magnitudes}.

\section{Convergence guarantees}
\label{sec:convergence_consequences}

Stationarity is averaged over all pre-update parameters, including the initial parameter and excluding the final post-update parameter. Both capped methods use the common maximum response length $T$ as their deterministic denominator. Expectations are along the run of the algorithm under consideration.

\subsection{Convergence with independent prompt sampling}
\label{sec:batch-convergence}

Independent prompt draws and empirical group covariance yield separate $B$ and $|G|$ contributions, with explicit coefficients from the directional score second moment and its trace, respectively. Dependence between response groups is permitted, so response batch size $B|G|$ cannot replace both sample counts in the variance bound.

\begin{theorem}[TIC-GRPO and GRPO$_2$ convergence, informal]
\label{thm:batch-convergence-informal}
Under Assumptions~\ref{L_smooth}--\ref{bounded_reward} and the conditional sampling protocol, let $N\geq2$, $K\geq1$, $B\geq1$, $|G|\geq2$, and $|\mathcal V|\geq2$. With separately selected admissible constant steps and fixed $R,L,c_+,|\mathcal V|$, the average squared gradient norm
\[
 \frac1{(N-1)K}\sum_{n=1}^{N-1}\sum_{k=0}^{K-1}
       \Expect\|\nabla J(\theta_{n,k})\|^2
\]
is bounded by the following respective orders:
\begin{align}
 \mathrm{TIC}:\quad&
 \mathcal{O}\!\left(\frac{T}{N-1}\right)
 +\mathcal{O}\!\left(T\sqrt{\frac{B^{-1}+|G|^{-1}}{N-1}}\right),
 \label{eq:tic-order}\\
 \mathrm{GRPO}_2:\quad&
 \mathcal{O}\!\left(\frac{T}{N-1}\right)
 +\mathcal{O}\!\left(T^{3/2}\sqrt{\frac{B^{-1}+|G|^{-1}}{N-1}}\right).
 \label{eq:grpo2-order}
\end{align}
The constants are independent of $T,N,K,B,$ and $|G|$. Expectations follow each algorithm's own run, and both guarantees include a zero initial gap.
\end{theorem}
Theorem~\ref{thm:batch-tic-convergence} in Appendix~\ref{app:batch-convergence-proof} gives the finite-step inequalities, full parameter dependence, admissible step-size interval, and separate step choices.

After separate step-size optimization, the sampling-term horizon coefficients are $T$ for TIC and $T^{3/2}$ for the token-level guarantee, a $\sqrt T$ improvement; the common finite-iteration term is linear in $T$. This compares upper guarantees, not performance lower bounds. It holds for the same $R,L,c_+,K,B,|G|,|\mathcal V|$ and an initial gap bounded independently of $T$, with $J^*-J(\theta_{1,0})\leq2R$. The explicit bounds track all dependencies; any variation of batch or vocabulary size with $T$ must also be included.

The square-root term determines the horizon comparison when it dominates the common finite-iteration term. Both methods have the same update at $K=1$, because every update is on policy; the two displayed bounds therefore represent different upper guarantees for that case. They also coincide at $T=1$. For $|G|=1$, all empirical advantages vanish and the universal bound $R^2LT$ applies. The cases $R=0$, $L=0$, or $|\mathcal V|=1$ have zero objective gradient. The step-size interval in Theorem~\ref{thm:batch-tic-convergence} is sufficient for the block ascent argument.

\subsection{Per-response normalization}

For original GRPO, the joint old-policy response-length variance is
\begin{equation}
 \sigma_{\theta_{n,0}}^2
 :=\operatorname{Var}_{c\sim p,\,s_T\sim\Pro_{\theta_{n,0}}(\cdot\mid c)}(|s_T|).
 \label{sigma}
\end{equation}
Its per-response denominator introduces a length-covariance contribution to the population update.

\begin{theorem}[GRPO convergence, informal]\label{thm:grpo-stationarity-informal}
Under the same sampling protocol and regularity assumptions, let
$N\geq2$, $K\geq1$, $|G|\geq2$, $\eta>0$, and
$\epsilon_{\low},\epsilon_{\high}>0$. For $K\geq2$, original GRPO satisfies
\begin{align}
 &\frac1{(N-1)K}\sum_{n=1}^{N-1}\sum_{k=0}^{K-1}
       \Expect\|\nabla J(\theta_{n,k})\|^2\leq \mathcal{O}\!\left(\frac{T}{\eta(N-1)}\right)
       +\mathcal{O}\!\left(\frac{\eta T^2}{N-1}
                   \sum_{n=1}^{N-1}\Expect[\mathcal M_n^2]\right)\notag\\
 &\quad+\mathcal{O}\!\left(\eta T^{5/2}
           \sqrt{\frac1{N-1}\sum_{n=1}^{N-1}\Expect[\mathcal M_n^2]}\right)
       +\mathcal{O}\!\left(\frac{T}{N-1}\sum_{n=1}^{N-1}\Expect\!\left[
          \frac{\sigma_{\theta_{n,0}}^2}
               {\Expect_{\theta_{n,0}}[|s_T|]}\right]\right).
 \label{eq:grpo-order}
\end{align}
Constants depend only on $R,L,K,\epsilon_{\low},\epsilon_{\high},|\mathcal V|$ and
$J^*-J(\theta_{1,0})$. For $K=1$, suitable constant steps, including when the initial gap is zero, give
\begin{align}
 \frac1{N-1}\sum_{n=1}^{N-1}\Expect\|\nabla J(\theta_{n,0})\|^2
 &\leq \mathcal{O}\!\left(\frac{T^{3/2}}{\sqrt{N-1}}\right)
       +\mathcal{O}\!\left(\frac{T}{N-1}\sum_{n=1}^{N-1}\Expect\!\left[
          \frac{\sigma_{\theta_{n,0}}^2}
               {\Expect_{\theta_{n,0}}[|s_T|]}\right]\right).
 \label{eq:grpo-on-policy-order}
\end{align}
\end{theorem}
Theorem~\ref{thm:grpo-stationarity} in Appendix~\ref{su_grpo} gives the explicit inequalities and step choices. For a positive initial gap, the $K=1$ choice minimizes the two explicit optimization terms; the length term remains evaluated along the resulting run.

The individual-length term from per-response normalization is at most a constant times $T(\sqrt T-1)^2$, without a $B^{-1}$ or $|G|^{-1}$ factor. For general $K$, the inverse-probability moments in~\eqref{eq:grpo-order} describe the original clipping rule's upper tail; they remain inside expectations along the run at the selected step size. For $K=1$, all updates are on policy, giving~\eqref{eq:grpo-on-policy-order} directly from the score moment. All bounds can be capped by $R^2LT$, including when a clipping margin is zero and the ratio-one subgradient is unspecified.

\section{Experiments}\label{experiments_pre}
We evaluate TIC-GRPO on Qwen3-1.7B and Qwen3-8B~\citep{yang2025qwen3}
against GSPO~\citep{zheng2025group} and a DAPO implementation of the
critic-free group-relative GRPO baseline~\citep{yu2025dapo}. Table~\ref{tab:main_tab} shows that
TIC-GRPO is best on all four benchmarks; Appendix~\ref{sec:experiments_setup}--\ref{sec:experiments_3}
reports setup details, learning curves, and evaluations of the component variants.

\begin{table}[H]
\caption{Main results. Avg@1 uses greedy decoding and Avg@32 uses 32 samples; parentheses are gains over GRPO.}
\label{tab:main_tab}
\centering
\begingroup
\setlength{\tabcolsep}{2.0pt}
\renewcommand{\arraystretch}{1.02}

\begin{tabular*}{\columnwidth}{@{\extracolsep{\fill}}lcccc@{}}
\toprule
\multirow{2}{*}{\textbf{Method}}
& \textbf{AIME24} & \textbf{AIME25} & \textbf{MATH500} & \textbf{Live-CodeBench} \\
& (Avg@32) & (Avg@32) & (Avg@1) & (Avg@1) \\
\specialrule{0.45pt}{0pt}{0pt}

\multicolumn{5}{c}{\textbf{Qwen3-1.7B}} \\
\midrule
GRPO      & 9.17 & 5.31 & 66.6 & 10.67 \\
GSPO      & 10.31 (+1.14) & 6.24 (+0.93) & 69.0 (+2.4) & 11.23 (+0.56) \\
\textbf{TIC-GRPO}
          & \textbf{11.77 (+2.60)} & \textbf{6.98 (+1.67)} & \textbf{69.8 (+3.2)} & \textbf{12.11 (+1.44)} \\
\specialrule{0.45pt}{0pt}{0pt}

\multicolumn{5}{c}{\textbf{Qwen3-8B}} \\
\midrule
GRPO      & 31.35 & 22.9 & 88.6 & 19.88 \\
GSPO      & 30.21 (-1.14) & 22.5 (-0.4) & 88.4 (-0.2) & 20.00 (+0.12) \\
\textbf{TIC-GRPO}
          & \textbf{33.34 (+1.99)} & \textbf{24.12 (+1.22)} & \textbf{90.0 (+1.4)} & \textbf{21.00 (+1.12)} \\
\bottomrule
\end{tabular*}
\endgroup
\end{table}

\section{Conclusion}
TIC-GRPO combines upper-tail clipping with trajectory-level importance correction in a critic-free group-relative update. Under common maximum-length normalization, matching upper and lower bounds give worst-case response-level variance orders $1$ for GRPO$_2$ and $1/T$ for TIC-GRPO. A full-trajectory change of measure preserves the score-martingale second moment through the empirical-baseline and block-convergence analysis. With separately optimized step sizes, the sampling term in TIC's stationarity bound has a linear horizon coefficient, improving by $\sqrt T$ over the displayed token-level GRPO$_2$ guarantee. The bounds separate prompt and within-group sampling. Experiments and ablations evaluate both mechanisms in sampled-token-normalized training on reasoning and coding benchmarks.

\begin{ack}
The authors received no third-party funding or in-kind support for this work. The authors declare no competing financial interests.
\end{ack}

\bibliographystyle{plainnat}

\clearpage
\appendix
\begingroup
\raggedbottom
\setlength{\intextsep}{6pt}
\section{Experiments}\label{sec:experiments}

\subsection{Experimental setup}\label{sec:experiments_setup}

\textbf{Models and Training Data.} We conduct all experiments using the Qwen3~\citep{yang2025qwen3} family of models, including Qwen3-0.6B, Qwen3-1.7B, and Qwen3-8B. The training data consist of a hybrid dataset that combines the full DAPO-Math-17K~\citep{yu2025dapo} corpus with a subset of the AIME benchmark (1983--2022), resulting in several hundred additional high-difficulty mathematical reasoning problems. All RL training is conducted within the VeRL~\citep{sheng2024hybridflow} framework, and all models are trained for a single epoch, ensuring that each prompt is used exactly once during training. We use a DAPO-style implementation as the experimental GRPO baseline and implement the proposed variants within the same training framework. DAPO combines the critic-free group-relative update with asymmetric clipping and a token-level loss~\citep{yu2025dapo}. The experimental implementation divides the capped update numerator by the sampled mean number of active response tokens. The theoretical results concern the maximum-length-normalized updates and fresh conditional sampling; the experiments evaluate the clipping and trajectory-correction mechanisms with sampled-token normalization and trajectory reuse.

\textbf{Evaluation.} We evaluate on three mathematical reasoning benchmarks and one coding benchmark: MATH500, the 500-problem subset of MATH introduced by \citet{lightman2023lets} from the dataset of \citet{hendrycks2021measuring}; the 2024 and 2025 American Invitational Mathematics Examinations (AIME24 and AIME25)~\citep{maaAIME}; and LiveCodeBench~\citep{jain2024livecodebench}. We report Avg@1 on MATH500 and LiveCodeBench. For AIME24 and AIME25, which each contain 30 problems, we report Avg@32.

\begin{itemize}
    \item \textbf{Avg@1:} Standard accuracy computed using greedy decoding, measuring deterministic single-response performance.
    \item \textbf{Avg@32:} Average accuracy over 32 sampled responses per problem, obtained with a temperature of 1.0 and top-$p$ of 1.0. This metric estimates mean accuracy under the specified sampling distribution.
\end{itemize}

\textbf{Training Configuration.} We adopt a total batch size of 128 with a mini-batch size of 32. Each sampled trajectory is reused for four gradient updates before the old policy is refreshed. This configuration enables efficient reuse of collected samples while maintaining stable optimization. Training of Qwen3-1.7B is performed on a single H200 GPU for approximately 24 hours, whereas Qwen3-8B is trained on two H200 nodes for 48 hours. For the larger model, gradient accumulation is applied to ensure that the effective global batch size remains consistent with that of Qwen3-1.7B. Furthermore, to eliminate confounding factors introduced by stochastic data processing, we disable both Dynamic Sampling and Soft Overlong Punishment in all experiments. Unless otherwise specified, all reported results follow this unified experimental configuration.

\subsection{Primary empirical results of TIC-GRPO}\label{sec:experiments_2}

As concluded in Sec.~\ref{experiments_pre}, our proposed TIC-GRPO consistently outperforms recent advanced GRPO-style baselines~\citep{yu2025dapo, zheng2025group} across different model scales. In this section, we further present the training reward and accuracy curves corresponding to the results in Table~\ref{tab:main_tab}, together with additional experimental evidence that provides a more comprehensive evaluation of TIC-GRPO.

Figure~\ref{fig:prime_exp} summarizes the primary empirical results on both Qwen3-1.7B and Qwen3-8B models, where each row corresponds to a different model size. For each model, we report the AIME24 Avg@32 accuracy curve (left) alongside the corresponding training reward curve (right), showing the evolution of evaluation accuracy and training reward.

TIC-GRPO attains the largest reported AIME24 scores in Table~\ref{tab:main_tab} at both model scales. Figure~\ref{fig:prime_exp} shows its evaluation accuracy and training reward over the course of optimization. The accuracy curves cross during training, and the relative ordering of training rewards can differ from that of evaluation accuracy.

\begin{figure}[H]

    \centering
    \begin{minipage}{0.95\linewidth}
        \centering

        \begin{subfigure}{0.48\linewidth}
            \centering
            \includegraphics[width=\linewidth]{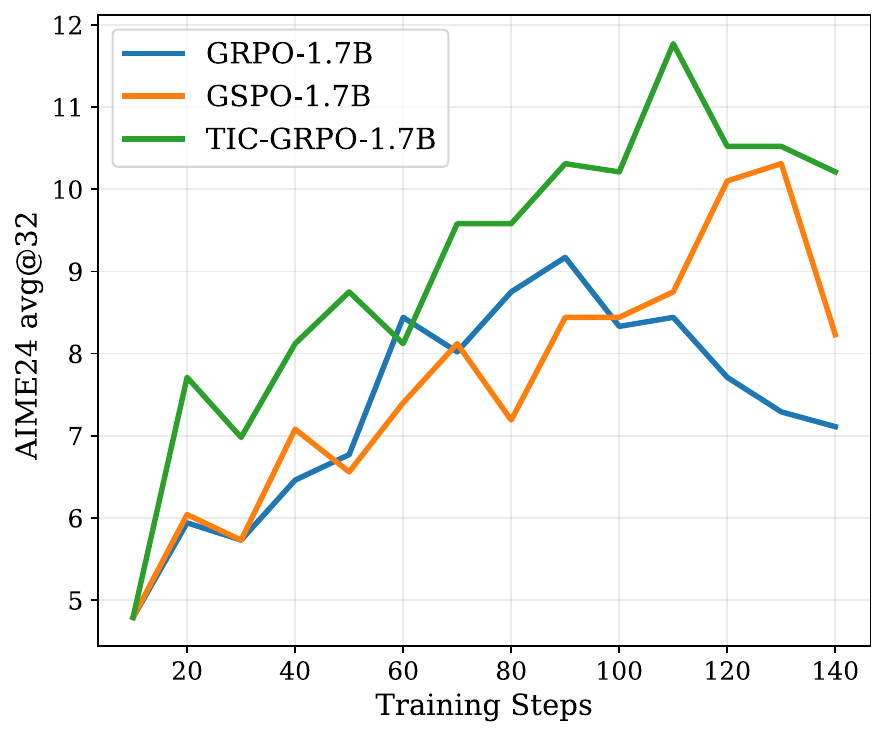}
            \caption{AIME24 Avg@32 accuracy on Qwen3-1.7B.}
            \label{fig:sub1'}
        \end{subfigure}
        \hfill
        \begin{subfigure}{0.48\linewidth}
            \centering
            \includegraphics[width=\linewidth]{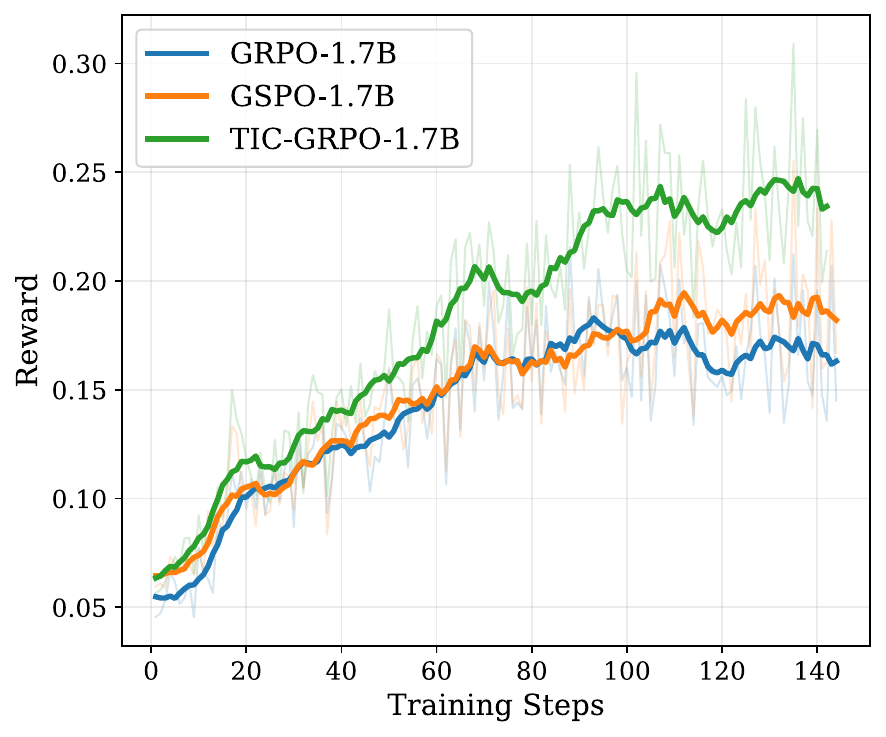}
            \caption{Training reward curves on Qwen3-1.7B.}
            \label{fig:sub2'}
        \end{subfigure}

        \begin{subfigure}{0.48\linewidth}
            \centering
            \includegraphics[width=\linewidth]{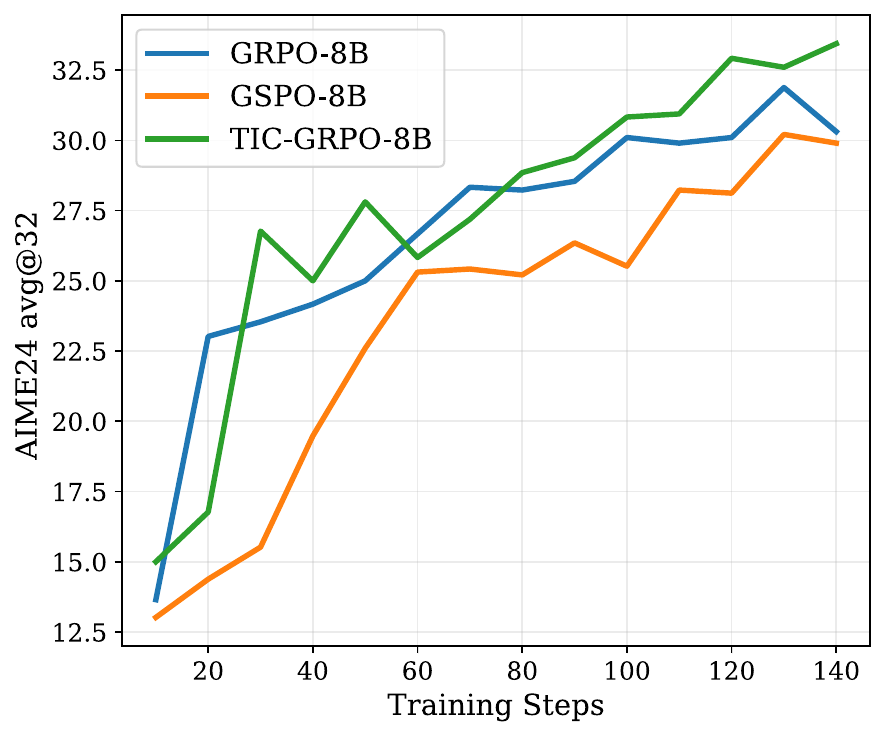}
            \caption{AIME24 Avg@32 accuracy on Qwen3-8B.}
            \label{fig:sub3'}
        \end{subfigure}
        \hfill
        \begin{subfigure}{0.48\linewidth}
            \centering
            \includegraphics[width=\linewidth]{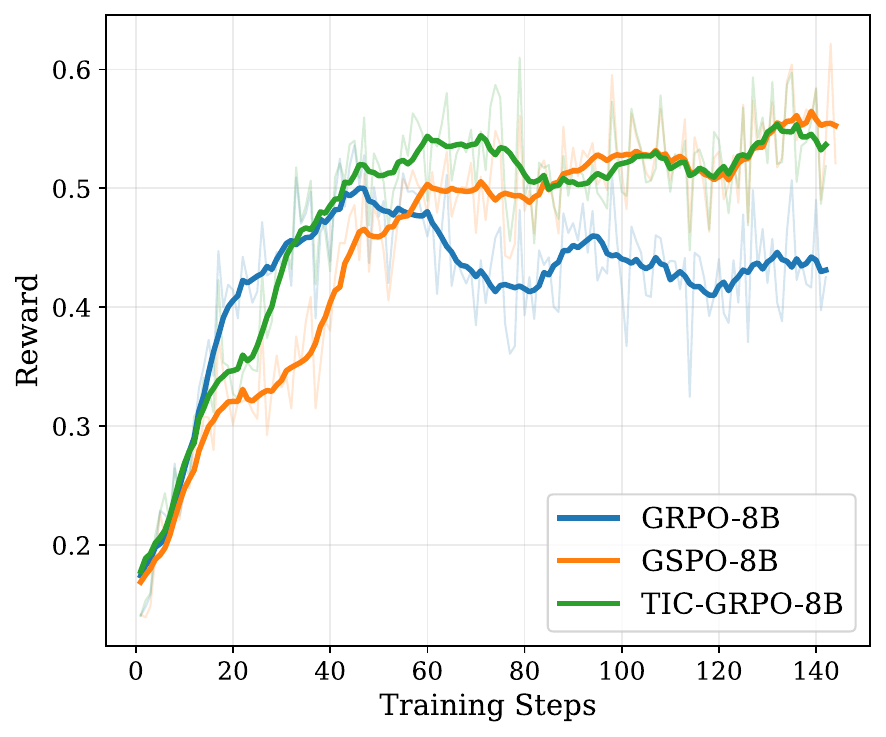}
            \caption{Training reward curves on Qwen3-8B.}
            \label{fig:sub4'}
        \end{subfigure}
    \end{minipage}

    \caption{Training dynamics of different GRPO variants on Qwen3-1.7B and Qwen3-8B models. Left panels show the AIME24 Avg@32 accuracy curves, while right panels report the corresponding training reward curves. Each panel compares the three methods at the recorded training steps.
}
    \label{fig:prime_exp}
\end{figure}

\subsection{Ablation study}\label{sec:experiments_3}

We evaluate component variants of TIC-GRPO on Qwen3-1.7B, using the experimental GRPO baseline for comparison. The variants use the mechanisms introduced in Section~\ref{sec:tic_grpo}:

\begin{itemize}
    \item \textbf{GRPO+UB\_Clip}: uses the proposed \textbf{up-only clipping mechanism} with sampled-token normalization. It retains separate token-level importance weights and caps their upper tails for either sign of the group-relative advantage. This variant evaluates upper-tail control while retaining token-level weighting.
    \item \textbf{GRPO+Traj\_IS}: replaces the token-level importance sampling used in GRPO with \textbf{trajectory-level importance sampling}, while keeping the original optimization procedure unchanged. The full-response likelihood ratio supplies a shared importance weight to the active-token scores within a response, allowing us to assess trajectory-level correction separately from up-only clipping.
    \item \textbf{TIC-GRPO}: combines up-only clipping with trajectory-level importance sampling. The upper cap acts on the shared trajectory weight, so this variant brings both components into the same update. Its comparison with each individual variant measures the empirical benefit of their combination.
\end{itemize}

\begin{table}[H]
\caption{Ablation study of TIC-GRPO on Qwen3-1.7B. We compare the GRPO baseline with two component variants---up-only clipping (GRPO+UB\_Clip) and trajectory-level importance sampling (GRPO+Traj\_IS)---as well as their combination (TIC-GRPO). }
\label{tab:main_tab'}
\centering
\setlength{\tabcolsep}{5pt}
\renewcommand{\arraystretch}{1.15}

\begin{tabular}{lcccc}
\toprule
\multirow{2}{*}{\textbf{Method}}
& \textbf{AIME24} & \textbf{AIME25} & \textbf{MATH500} & \textbf{Live-CodeBench} \\
& (Avg@32) & (Avg@32) & (Avg@1) & (Avg@1) \\
\specialrule{0.8pt}{0pt}{0pt}

\multicolumn{5}{c}{\textbf{Qwen3-1.7B}} \\
\midrule
GRPO
& 9.17 & 5.31 & 66.6 & 10.67 \\

GRPO+UB\_Clip
& 10.31 (+1.14) & 6.64 (+1.33) & 67.4 (+0.8) & 11.34 (+0.67) \\

GRPO+Traj\_IS
& 10.62 (+1.45) & 6.77 (+1.46) & 68.0 (+1.4) & 11.44 (+0.77) \\

\textbf{TIC-GRPO}
& \textbf{11.77 (+2.60)} & \textbf{6.98 (+1.67)} & \textbf{69.8 (+3.2)} & \textbf{12.11 (+1.44)} \\
\bottomrule
\end{tabular}
\end{table}

\begin{figure}[H]
    \centering
    \begin{minipage}{0.90\linewidth}
        \centering
        \begin{subfigure}[t]{0.48\linewidth}
            \centering
            \vspace{0pt}
            \includegraphics[height=0.83\linewidth]{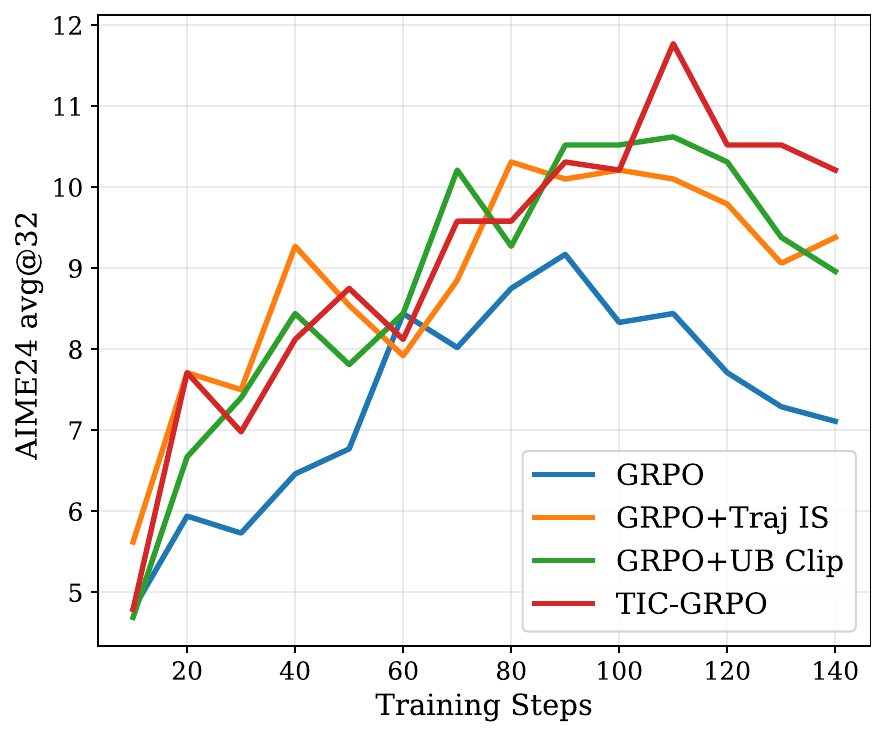}
            \caption{AIME24 Avg@32 accuracy curves during training.}
        \end{subfigure}
        \hfill
        \begin{subfigure}[t]{0.48\linewidth}
            \centering
            \vspace{0pt}
            \includegraphics[height=0.83\linewidth]{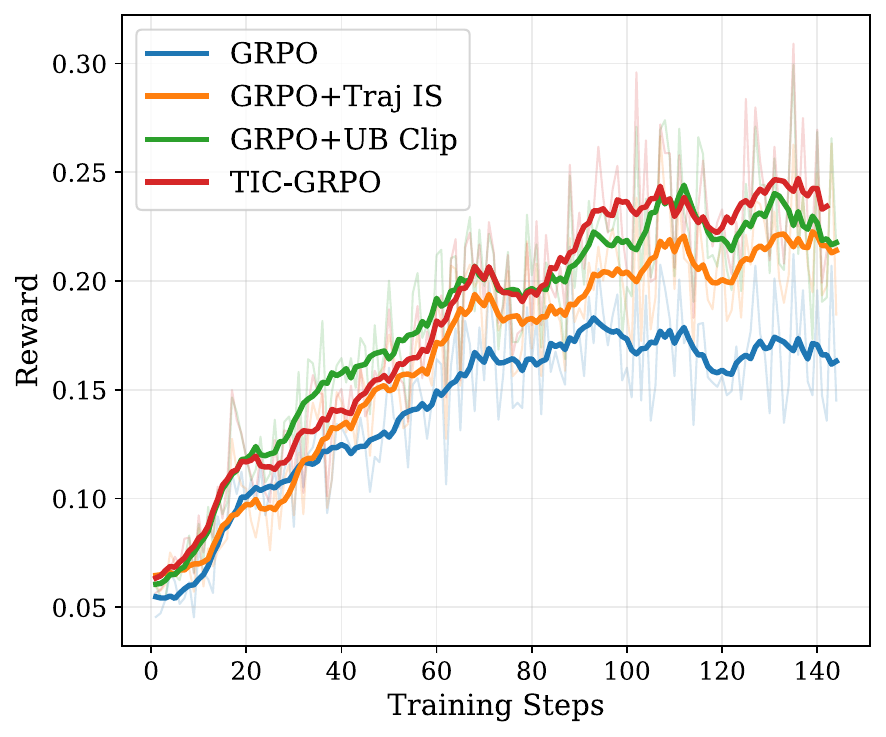}
            \caption{Corresponding training reward curves.}
            \label{fig:sub2''}
        \end{subfigure}
    \end{minipage}
    \caption{Ablation analysis of TIC-GRPO on Qwen3-1.7B.}
    \label{fig:ablation}
\end{figure}

Table~\ref{tab:main_tab'} reports the benchmark scores, and Figure~\ref{fig:ablation} shows the corresponding AIME24 accuracy and training reward curves. Both component variants improve on the GRPO baseline across all four benchmarks. The trajectory-level variant exceeds the up-only clipping variant, attaining scores of 10.62, 6.77, 68.0, and 11.44 on AIME24, AIME25, MATH500, and LiveCodeBench, respectively. Combining the two components raises these scores to 11.77, 6.98, 69.8, and 12.11. Relative to the stronger individual variant, TIC-GRPO therefore gains 1.15 points on AIME24, 0.21 on AIME25, 1.8 on MATH500, and 0.67 on LiveCodeBench. The combined method leads on both mathematical reasoning and coding evaluations, supporting the complementary practical effects of upper-tail clipping and trajectory-level importance correction.

\subsection{Update magnitudes including GSPO}
\label{sec:gspo-experiments}

We compare the DAPO-style GRPO baseline, GRPO$_2$, GSPO, and TIC-GRPO on Qwen3-0.6B and Qwen3-1.7B using matched data, optimizer, and training settings as described in Appendix~\ref{sec:experiments_setup}. At each of the first 200 optimizer steps, we record $\|\theta_{k+1}-\theta_k\|_2^2$, the squared Euclidean distance between the parameter iterates immediately before and after that step. Table~\ref{tab:gspo-update-magnitudes} reports its mean within each of eight consecutive 25-step intervals, followed by its mean and temporal standard deviation across all 200 steps. The standard deviation summarizes variation across optimizer steps within a run.

The interval summaries follow the optimization process: the first row captures initial updates, while successive rows track how update magnitudes evolve as the policy changes. Because all intervals contain the same number of steps, they contribute equally to the overall mean. The overall temporal standard deviation is computed from individual steps and therefore also reflects fluctuations within each interval. We compare methods within each model scale under the matched setup. Together, the mean and standard deviation describe the typical squared parameter displacement and the stability of its magnitude over the training window.

\begin{table}[H]
\centering
\caption{Squared Euclidean distances between consecutive parameter iterates, in units of $10^6$. Interval rows report temporal means; each overall row reports the mean $\pm$ temporal standard deviation across optimizer steps 1--200.}
\label{tab:gspo-update-magnitudes}
\begin{tabular}{lrrrr}
\toprule
Optimizer steps & GRPO & GRPO$_2$ & GSPO & TIC-GRPO \\
\midrule
\multicolumn{5}{c}{\textbf{Qwen3-0.6B}} \\
\midrule
1--25    & 26.33 & 31.63 & 26.93 & 23.54 \\
26--50   & 45.29 & 52.39 & 45.02 & 29.09 \\
51--75   & 48.37 & 40.25 & 41.40 & 28.02 \\
76--100  & 32.09 & 31.88 & 31.81 & 26.54 \\
101--125 & 30.70 & 30.42 & 28.91 & 25.50 \\
126--150 & 32.51 & 25.96 & 33.17 & 25.63 \\
151--175 & 25.46 & 25.76 & 28.77 & 26.06 \\
176--200 & 29.73 & 30.17 & 30.39 & 25.97 \\
\midrule
Overall & $33.81\pm11.14$ & $33.56\pm10.66$ & $33.30\pm9.03$ & $26.29\pm2.78$ \\
\midrule
\multicolumn{5}{c}{\textbf{Qwen3-1.7B}} \\
\midrule
1--25    & 88.29 & 85.75 & 98.23 & 78.71 \\
26--50   & 143.71 & 172.75 & 150.52 & 91.61 \\
51--75   & 139.63 & 163.99 & 168.42 & 84.64 \\
76--100  & 128.41 & 118.50 & 117.31 & 81.83 \\
101--125 & 117.28 & 88.70 & 88.94 & 77.17 \\
126--150 & 102.50 & 87.26 & 81.86 & 78.69 \\
151--175 & 94.98 & 95.76 & 76.61 & 76.94 \\
176--200 & 94.46 & 86.62 & 92.60 & 75.63 \\
\midrule
Overall & $113.66\pm26.57$ & $112.41\pm37.90$ & $109.31\pm37.18$ & $80.65\pm8.86$ \\
\bottomrule
\end{tabular}
\end{table}

TIC-GRPO has the smallest overall mean and temporal standard deviation among the four methods at both model scales. Relative to GSPO, its mean is lower by about 21\% on Qwen3-0.6B and 26\% on Qwen3-1.7B, while its temporal standard deviation is lower by about 69\% and 76\%, respectively. The interval means also show the smaller update magnitudes during steps 26--75: GSPO reaches 45.02 and 41.40 on Qwen3-0.6B, versus 29.09 and 28.02 for TIC-GRPO; on Qwen3-1.7B, the corresponding values are 150.52 and 168.42 for GSPO, versus 91.61 and 84.64 for TIC-GRPO.

\section{Score moments and objective geometry}
\label{app:score_geometry}

\begin{lemma}[Score moments]\label{lem:score-moments}
Under Assumption~\ref{L_smooth}, at every active prefix,
\begin{align}
 \|g_t^\theta\|^2&\leq-2L\log\Pro_\theta(s_t\mid s_{t-1}),\label{eq:score-self-bound}\\
 \left\|\Expect_\theta[g_t^\theta(g_t^\theta)^\top\mid s_{t-1}]\right\|_{\mathrm{op}}
 &\leq L,\qquad
 \Expect_\theta[\|g_t^\theta\|^2\mid s_{t-1}]
 \leq2L\log|\mathcal V|.\label{eq:token-fisher}
\end{align}
Consequently, for each prompt,
\begin{equation}
\label{eq:trajectory-fisher}
\left\|\Expect_\theta[G_T^\theta(G_T^\theta)^\top\mid c]\right\|_{\mathrm{op}}
 \leq L\Expect_\theta[|s_T|\mid c],\qquad \Expect_\theta[\|G_T^\theta\|^2\mid c]
 \leq2L\log|\mathcal V|\Expect_\theta[|s_T|\mid c].
\end{equation}
\end{lemma}
\begin{proof}
The negative log probability is nonnegative and has an $L$-Lipschitz gradient on the parameter space. The descent inequality evaluated one gradient step of length $1/L$ away gives~\eqref{eq:score-self-bound} when $L>0$. If $L=0$, a globally affine nonnegative negative log probability has zero gradient, so every asserted inequality is immediate.

For the directional Fisher estimate, restrict the finite next-token probability vector to any line in parameter space. Its log derivatives are Lipschitz. Differentiating the normalization identity twice at the almost-everywhere points of second differentiability shows that the expected squared directional score is the negative expected directional second derivative of the log probability. The latter is bounded by $L$ times the squared norm of the direction. Continuity of the finite sum of squared scores extends this bound to every point of the line. This proves the operator-norm assertion. Taking the expectation in~\eqref{eq:score-self-bound} and using the entropy bound for a distribution on $|\mathcal V|$ symbols proves the trace assertion.

The conditional mean of $g_t^\theta$ is zero, since the derivatives of the next-token probabilities sum to zero. For $i<j$, $g_i^\theta$ is measurable before token $j$, so conditioning on that prefix gives
\[
 \Expect_\theta[g_i^\theta(g_j^\theta)^\top\mid c]=0.
\]
The active-token indicator is determined by the preceding prefix. Thus the diagonal terms in the second moment of the stopped sum are bounded by the one-token estimates times the expected number of active transitions. This proves both inequalities in Eq.~\eqref{eq:trajectory-fisher} and Property~\ref{prop:martingale_identity}. The action that terminates the response is included; only subsequent padding transitions have zero score.
\end{proof}

\begin{lemma}[Stopped normalized score]\label{lem:stopped-score}
For each prompt,
\begin{equation}
 \left\|\Expect_\theta\left[
 \frac{G_T^\theta(G_T^\theta)^\top}{|s_T|^2}\,\middle|\,c\right]\right\|_{\mathrm{op}}
 \leq L\sum_{t=1}^{T}\frac1{t^2}.
 \label{eq:stopped-second}
\end{equation}
Its trace is at most $2L\log|\mathcal V|\sum_{t=1}^{T}t^{-2}$.
\end{lemma}
\begin{proof}
Telescope the squared projection of the partial score sum divided by the square of its time index, stopping at $|s_T|$. The mixed term with the next increment has zero conditional mean. The coefficient of the preceding squared projection is $t^{-2}-(t-1)^{-2}\leq0$ for $t\geq2$. After treating $t=1$ separately, the directional conditional bound~\eqref{eq:token-fisher} gives~\eqref{eq:stopped-second}. The event that token $t$ is active is measurable before that token is sampled, so the stopping indicators preserve the conditional calculation. The same telescoping argument applied to the squared norm and the trace bound proves the trace assertion.
\end{proof}

\begin{lemma}[Reward geometry]\label{lem:objective-geometry}
For all parameters $\theta,\theta'$,
\begin{equation}
\label{eq:gradient-envelope}
\|\nabla J(\theta)\|^2\leq R^2L\Expect_\theta[|s_T|]\leq R^2LT,\qquad \|\nabla J(\theta')-\nabla J(\theta)\|\leq2RLT\|\theta'-\theta\|.
\end{equation}
\end{lemma}
\begin{proof}
The policy-gradient identity and the directional form of the first inequality in Eq.~\eqref{eq:trajectory-fisher} give the first bound in Eq.~\eqref{eq:gradient-envelope} by Cauchy--Schwarz, using $|r|\leq R$ and then averaging over prompts.

The trajectory score is $L|s_T|$-Lipschitz. Along any line segment, differentiating the reward gradient almost everywhere gives the expected reward times the trajectory-score derivative plus the expected reward times the outer product of the trajectory score applied to the parameter direction. Their expected absolute projections onto a unit direction are bounded, respectively, by $RL\Expect_\theta|s_T|$ and $RL\Expect_\theta|s_T|$, times the norm of the parameter direction. The first uses score Lipschitzness; the second uses directional Fisher and Cauchy--Schwarz. Integration proves the second bound in Eq.~\eqref{eq:gradient-envelope}. Finite response supports justify the calculation for each prompt, and the uniform bound $2RLT$ permits averaging over prompts. Derivatives along the segment suffice; ambient second differentiability at every point is not required.
\end{proof}

\section{Empirical covariance and block convergence}
\label{app:batch-convergence-proof}

\begin{theorem}[TIC-GRPO and GRPO$_2$ convergence]
\label{thm:batch-tic-convergence}\label{thm:batch-grpo2-convergence}
Under Assumptions~\ref{L_smooth} and~\ref{bounded_reward} and the conditional sampling protocol, let $N\geq2$, $K\geq1$, $B\geq1$, $|G|\geq2$, and $|\mathcal V|\geq2$. For $0<\eta\leq(128RLK)^{-1}$, TIC-GRPO satisfies
\begin{align}
 &\frac1{(N-1)K}\sum_{n=1}^{N-1}\sum_{k=0}^{K-1}
       \Expect\|\nabla J(\theta_{n,k})\|^2\notag\\
 &\leq\frac{24T[J^*-J(\theta_{1,0})]}{\eta K(N-1)}\notag\\
 &\quad+\left(1-\frac1{|G|}\right)^2 R^2Lc_+T
       \left(\frac1B+\frac{32\log|\mathcal V|}{|G|}\right)
       \left[48RLK\eta+392R^2L^2K(K-1)\eta^2\right].
 \label{eq:batch-tic-convergence}
\end{align}
Under the same conditions, GRPO$_2$ satisfies
\begin{align}
 &\frac1{(N-1)K}\sum_{n=1}^{N-1}\sum_{k=0}^{K-1}
       \Expect\|\nabla J(\theta_{n,k})\|^2\notag\\
 &\leq\frac{24T[J^*-J(\theta_{1,0})]}{\eta K(N-1)}\notag\\
 &\quad+\left(1-\frac1{|G|}\right)^2 R^2Lc_+T^2
       \left(\frac1B+\frac{32\log|\mathcal V|}{|G|}\right)
       \left[48RLK\eta+392R^2L^2K(K-1)\eta^2\right].
 \label{eq:batch-grpo2-convergence}
\end{align}
Each method selects its own constant step size. For a positive initial gap, minimizing its respective displayed bound over $0<\eta\leq(128RLK)^{-1}$ yields, for TIC-GRPO,
\begin{align}
 &\frac1{(N-1)K}\sum_{n=1}^{N-1}\sum_{k=0}^{K-1}
       \Expect\|\nabla J(\theta_{n,k})\|^2\notag\\
 &\leq\frac{6144R^2LT}{N-1}
 +102\left(1-\frac1{|G|}\right)R^2LT
       \sqrt{\frac{c_+}{N-1}
                   \left(\frac1B+\frac{32\log|\mathcal V|}{|G|}\right)},
 \label{eq:batch-tic-rate}
\end{align}
and, for GRPO$_2$,
\begin{align}
 &\frac1{(N-1)K}\sum_{n=1}^{N-1}\sum_{k=0}^{K-1}
       \Expect\|\nabla J(\theta_{n,k})\|^2\notag\\
 &\leq\frac{6144R^2LT}{N-1}
 +102\left(1-\frac1{|G|}\right)R^2LT
       \sqrt{\frac{c_+T}{N-1}
                   \left(\frac1B+\frac{32\log|\mathcal V|}{|G|}\right)}.
 \label{eq:batch-grpo2-rate}
\end{align}
The same two bounds hold for a zero initial gap with the explicit positive step choices given below.
\end{theorem}

All inner-step conditional expectations in this appendix are given $\mathscr F_{n,k}$. In particular, the current and old parameters are fixed before drawing the next batch. Token sums include only active transitions, and the normalizing denominator is the deterministic maximum length $T$.

\subsection{Moments of the capped score features}

For the TIC feature, the inequality $(\rho_{0:T}^+)^2\leq c_+\rho_{0:T}$ changes the entire response law. Lemma~\ref{lem:score-moments} then gives, for every prompt,
\begin{align}
 &\left\|\Expect_{\theta_{n,0}}\!\left[
   (\rho_{0:T}^+)^2G_T^{\theta_{n,k}}(G_T^{\theta_{n,k}})^\top
                       \mid c\right]\right\|_{\mathrm{op}}
       \leq c_+LT,\notag\\
 &\Expect_{\theta_{n,0}}\!\left[
   (\rho_{0:T}^+)^2\|G_T^{\theta_{n,k}}\|^2\mid c\right]
       \leq2c_+LT\log|\mathcal V|.
 \label{eq:tic-feature-moments}
\end{align}
The current-policy trajectory-score identity is applied after the change of measure. It applies to the score before multiplication by the reward or the empirical advantage.

For GRPO$_2$, Cauchy--Schwarz for the $T$ token positions gives the positive-semidefinite matrix inequality
\[
 \left(\sum_{t=1}^{T}\rho_t^+g_t^{\theta_{n,k}}\right)
 \left(\sum_{t=1}^{T}\rho_t^+g_t^{\theta_{n,k}}\right)^\top
 \preceq T\sum_{t=1}^{T}(\rho_t^+)^2
                  g_t^{\theta_{n,k}}(g_t^{\theta_{n,k}})^\top.
\]
At an active old-policy prefix, $(\rho_t^+)^2\leq c_+\rho_t$ changes only the next-token law. The directional bound in~\eqref{eq:token-fisher} gives an operator norm at most $c_+L$ for each term; its trace is at most $2c_+L\log|\mathcal V|$. Summing the $T$ positions proves
\begin{align}
 &\left\|\Expect_{\theta_{n,0}}\!\left[
 \left(\sum_{t=1}^{T}\rho_t^+g_t^{\theta_{n,k}}\right)
 \left(\sum_{t=1}^{T}\rho_t^+g_t^{\theta_{n,k}}\right)^\top
                          \mid c\right]\right\|_{\mathrm{op}}
      \leq c_+LT^2,\notag\\
 &\Expect_{\theta_{n,0}}\!\left[
       \left\|\sum_{t=1}^{T}\rho_t^+g_t^{\theta_{n,k}}\right\|^2
                              \mid c\right]
      \leq2c_+LT^2\log|\mathcal V|.
 \label{eq:grpo2-feature-moments}
\end{align}
This calculation bounds the complete token sum, including its cross terms, by a sum of single-token second moments.

\subsection{The empirical baseline and the batch variance}

The empirical covariance in a response group has expectation $1-|G|^{-1}$ times the population reward-feature covariance. To verify this, expand the empirical reward mean: among its $|G|^2$ products, $|G|$ contain a reward and feature from the same response, while the other $|G|(|G|-1)$ contain independent responses from that prompt. Consequently,
\begin{align}
 \Expect[\widehat G_{\mathrm{TIC}}^{(n,k)}\mid\mathscr F_{n,k}]
 &=\frac{1-|G|^{-1}}{T}\Expect_{\theta_{n,0}}\!\left[
    (r(s_T)-\mu_{\theta_{n,0},c})\rho_{0:T}^+
                                      G_T^{\theta_{n,k}}\right],\notag\\
 \Expect[\widehat G_{\mathrm{GRPO}_2}^{(n,k)}\mid\mathscr F_{n,k}]
 &=\frac{1-|G|^{-1}}{T}\Expect_{\theta_{n,0}}\!\left[
    (r(s_T)-\mu_{\theta_{n,0},c})
                  \sum_{t=1}^{T}\rho_t^+g_t^{\theta_{n,k}}\right].
 \label{eq:actual-group-means}
\end{align}
The expectation on the right includes the prompt draw.

Here is a finite-sample variance calculation retaining the same-group baseline. Write the empirical covariance as $1-|G|^{-1}$ times the average over unordered pairs of distinct responses of the product of the reward difference and the feature difference, divided by two. Each pair term has squared-norm expectation at most $4R^2$ times the single-response feature second moment: the reward difference is at most $2R$, and the expected squared feature difference is at most four times that moment. Disjoint pairs are independent. For a fixed pair, precisely $2|G|-3$ pairs, including itself, intersect it. Cauchy--Schwarz for the covariance of intersecting pairs and
\[
 \frac{2|G|-3}{\binom{|G|}{2}}\leq\frac4{|G|}
\]
therefore bound the group variance by $16(1-|G|^{-1})^2R^2/|G|$ times the single-response feature second moment. This also covers $|G|=2$.

The squared norm of each population reward-feature covariance is at most $R^2$ times the operator norm of the feature second-moment matrix. Indeed, project onto an arbitrary unit direction, apply Cauchy--Schwarz to the centered reward and the projected feature, and use $\operatorname{Var}(r\mid c)\leq R^2$. The bound is uniform over directions, so it bounds the norm of the mean vector without a trace factor.

Condition on the complete list of sampled prompt occurrences. The conditional mean of the batch is the average of the corresponding population group means. Independence of the $B$ prompt draws bounds the variance of that average by $B^{-1}$ times the preceding squared-mean bound. For the remaining within-group fluctuations, Jensen's inequality across the $B$ groups bounds the conditional squared norm of their average by the average of their conditional squared norms. The latter step permits arbitrary dependence between groups. The two fluctuations are orthogonal in expectation because the second has conditional mean zero given the prompt list. Applying the group calculation and~\eqref{eq:tic-feature-moments}--\eqref{eq:grpo2-feature-moments}, and dividing by $T^2$, yields
\begin{align}
 &\Expect\!\left[\left\|\widehat G_{\mathrm{TIC}}^{(n,k)}
       -\Expect[\widehat G_{\mathrm{TIC}}^{(n,k)}\mid\mathscr F_{n,k}]
                   \right\|^2\middle|\mathscr F_{n,k}\right]
 \leq\frac{(1-|G|^{-1})^2R^2c_+L}{T}
       \left(\frac1B+\frac{32\log|\mathcal V|}{|G|}\right),
 \label{eq:tic-batch-variance}\\
 &\Expect\!\left[\left\|\widehat G_{\mathrm{GRPO}_2}^{(n,k)}
       -\Expect[\widehat G_{\mathrm{GRPO}_2}^{(n,k)}\mid\mathscr F_{n,k}]
                   \right\|^2\middle|\mathscr F_{n,k}\right]
 \leq(1-|G|^{-1})^2R^2c_+L
       \left(\frac1B+\frac{32\log|\mathcal V|}{|G|}\right).
 \label{eq:grpo2-batch-variance}
\end{align}

\subsection{Stability of the actual mean update}

At $\theta_{n,k}=\theta_{n,0}$, both ratios equal one, so both expectations in~\eqref{eq:actual-group-means} equal $(1-|G|^{-1})\nabla J(\theta_{n,0})/T$. We bound the change of these complete population means along the parameter segment from $\theta_{n,0}$ to $\theta_{n,k}$, with the old response law and its reward mean held fixed.

For TIC, the derivative of the capped score feature has two terms: the capped ratio times the directional derivative of the trajectory score, and, where the ratio is below the cap, the ratio times the trajectory-score outer product applied to the parameter direction. For each fixed unit projection direction, the old-law expected absolute projection of the first term is bounded by $LT$ times the norm of the parameter direction. This uses the trajectory-score Lipschitz constant $L|s_T|$, the inequality $\rho_{0:T}^+\leq\rho_{0:T}$, and a full change of measure. The second term has the same bound, by the first inequality in Eq.~\eqref{eq:trajectory-fisher} and Cauchy--Schwarz under the current response law.

For GRPO$_2$, apply this derivative calculation separately at every active old-policy prefix. The first term, the capped ratio times the token-score derivative, has expected absolute projection at most $L$ times the norm of the parameter direction. In the second term, the ratio derivative is zero above the cap; below the cap, a next-token change of measure and~\eqref{eq:token-fisher} give the same bound. Summing at most $T$ active tokens bounds the expected absolute projection of the complete feature derivative by $2LT$ times that norm. Multiplication by $|r-\mu_{\theta_{n,0},c}|\leq2R$ is valid even though the reward depends on the entire response.

These derivative calculations hold almost everywhere along the segment. Lipschitz scores are absolutely continuous there; each probability and its capped ratio are locally absolutely continuous. On a level set where the ratio equals the cap, its segment derivative vanishes almost everywhere. Finite next-token and response supports justify each conditional differentiation. The bounds just obtained are uniform over prompts and along the segment, so integration and prompt averaging are legitimate. For either $\mathsf A\in\{\mathrm{TIC},\mathrm{GRPO}_2\}$ they prove
\begin{equation}
 \left\|\Expect[\widehat G_{\mathsf A}^{(n,k)}\mid\mathscr F_{n,k}]
       -\frac{1-|G|^{-1}}T\nabla J(\theta_{n,0})\right\|
 \leq4RL\|\theta_{n,k}-\theta_{n,0}\|.
 \label{eq:actual-mean-stability}
\end{equation}
This estimate holds also at $c_+=1$.

Decomposing a second moment into variance and squared mean, applying~\eqref{eq:actual-mean-stability}, and using $(1-|G|^{-1})^2\leq1$ gives
\begin{align}
 T\Expect[\|\widehat G_{\mathsf A}^{(n,k)}\|^2\mid\mathscr F_{n,k}]
 &\leq\frac4T\|\nabla J(\theta_{n,0})\|^2
      +32R^2L^2T\|\theta_{n,k}-\theta_{n,0}\|^2\notag\\
 &\quad+T\Expect\!\left[\left\|\widehat G_{\mathsf A}^{(n,k)}
    -\Expect[\widehat G_{\mathsf A}^{(n,k)}\mid\mathscr F_{n,k}]
                   \right\|^2\middle|\mathscr F_{n,k}\right].
 \label{eq:actual-relative-moment}
\end{align}

\subsection{Block ascent and all inner iterates}

For either method, the update recursion implies the pathwise bound
\[
 T\|\theta_{n,k}-\theta_{n,0}\|^2
 \leq\eta^2 k\sum_{j=0}^{k-1}T\|\widehat G_{\mathsf A}^{(n,j)}\|^2.
\]
Summing over $k$ and noting that each earlier moment has coefficient at most $K(K-1)/2$ gives
\begin{equation}
 \sum_{k=0}^{K-1}\Expect[T\|\theta_{n,k}-\theta_{n,0}\|^2]
 \leq\eta^2\frac{K(K-1)}2
              \sum_{k=0}^{K-1}\Expect[T\|\widehat G_{\mathsf A}^{(n,k)}\|^2].
 \label{eq:batch-displacement-closed}
\end{equation}
Since $RLK\eta\leq1/128$, the coefficient
$32R^2L^2\eta^2K(K-1)/2$ is at most $1/1024$. Substitute~\eqref{eq:batch-displacement-closed} into the sum of~\eqref{eq:actual-relative-moment} and absorb the displacement contribution on the left. The coefficient remaining on the left is at least $1-1/1024$, whose reciprocal is at most two. The result is
\begin{align}
 &\sum_{k=0}^{K-1}\Expect[T\|\widehat G_{\mathsf A}^{(n,k)}\|^2]
 \leq\frac{8K}{T}\Expect\|\nabla J(\theta_{n,0})\|^2\notag\\
 &\qquad+2T\sum_{k=0}^{K-1}\Expect\left\|\widehat G_{\mathsf A}^{(n,k)}
       -\Expect[\widehat G_{\mathsf A}^{(n,k)}\mid\mathscr F_{n,k}]
                                  \right\|^2.
 \label{eq:batch-moment-closed}
\end{align}
These quantities are finite by the uniform feature second moments and the finite update recursion.

For $|G|\geq2$, Young's inequality and~\eqref{eq:actual-mean-stability} give
\begin{align}
 &\left\langle\nabla J(\theta_{n,0}),
     \Expect[\widehat G_{\mathsf A}^{(n,k)}\mid\mathscr F_{n,k}]
                                                   \right\rangle\notag\\
 &\quad\geq\frac{\|\nabla J(\theta_{n,0})\|^2}{4T}
             -16R^2L^2T\|\theta_{n,k}-\theta_{n,0}\|^2.
 \label{eq:batch-mean-alignment}
\end{align}
In detail, the leading coefficient is at least $1/(2T)$, and
$4RL\|\nabla J(\theta_{n,0})\|\|\theta_{n,k}-\theta_{n,0}\|$
is at most $\|\nabla J(\theta_{n,0})\|^2/(4T)+16R^2L^2T\|\theta_{n,k}-\theta_{n,0}\|^2$.

Apply the $2RLT$ smoothness of $J$ to the complete block update. Cauchy--Schwarz across its $K$ steps, followed by~\eqref{eq:batch-mean-alignment} and~\eqref{eq:batch-displacement-closed}, yields
\begin{align}
 &\Expect[J(\theta_{n+1,0})-J(\theta_{n,0})]\notag\\
 &\geq\frac{\eta K}{4T}\Expect\|\nabla J(\theta_{n,0})\|^2
 -\eta\left[RLK\eta+8R^2L^2K(K-1)\eta^2\right]
            \sum_{k=0}^{K-1}\Expect[T\|\widehat G_{\mathsf A}^{(n,k)}\|^2].
 \label{eq:batch-block-ascent}
\end{align}
The bracket is at most $1/64$. Inserting~\eqref{eq:batch-moment-closed} leaves a gradient coefficient at least $\eta K/(8T)$. Sum over outer blocks and telescope $J$, using $J(\theta_{N,0})\leq J^*$. This proves
\begin{align}
 &\frac1{N-1}\sum_{n=1}^{N-1}
                   \frac{\Expect\|\nabla J(\theta_{n,0})\|^2}{T}
 \leq\frac{8[J^*-J(\theta_{1,0})]}{\eta K(N-1)}\notag\\
 &\quad+\frac{16T[RLK\eta+8R^2L^2K(K-1)\eta^2]}{(N-1)K}
   \sum_{n=1}^{N-1}\sum_{k=0}^{K-1}
    \Expect\left\|\widehat G_{\mathsf A}^{(n,k)}
     -\Expect[\widehat G_{\mathsf A}^{(n,k)}\mid\mathscr F_{n,k}]
                                        \right\|^2.
 \label{eq:batch-anchor-bound}
\end{align}

Finally, objective-gradient Lipschitzness implies
\[
 \|\nabla J(\theta_{n,k})\|^2
 \leq2\|\nabla J(\theta_{n,0})\|^2
       +8R^2L^2T^2\|\theta_{n,k}-\theta_{n,0}\|^2.
\]
Using~\eqref{eq:batch-displacement-closed} and~\eqref{eq:batch-moment-closed} to sum the second term gives
\begin{align*}
 &\frac1{(N-1)K}\sum_{n=1}^{N-1}\sum_{k=0}^{K-1}
                 \Expect\|\nabla J(\theta_{n,k})\|^2\\
 &\leq\frac{2+32R^2L^2K(K-1)\eta^2}{N-1}
             \sum_{n=1}^{N-1}\Expect\|\nabla J(\theta_{n,0})\|^2\\
 &\quad+\frac{8R^2L^2T^2K(K-1)\eta^2}{(N-1)K}
     \sum_{n=1}^{N-1}\sum_{k=0}^{K-1}
      \Expect\left\|\widehat G_{\mathsf A}^{(n,k)}
       -\Expect[\widehat G_{\mathsf A}^{(n,k)}\mid\mathscr F_{n,k}]
                    \right\|^2.
\end{align*}
The anchor-gradient coefficient $2+32R^2L^2K(K-1)\eta^2$ is at most three. Substitution of~\eqref{eq:batch-anchor-bound} therefore gives
\begin{align}
 &\frac1{(N-1)K}\sum_{n=1}^{N-1}\sum_{k=0}^{K-1}
                    \Expect\|\nabla J(\theta_{n,k})\|^2\notag\\
 &\leq\frac{24T[J^*-J(\theta_{1,0})]}{\eta K(N-1)}\notag\\
 &\quad+\frac{T^2[48RLK\eta+392R^2L^2K(K-1)\eta^2]}{(N-1)K}
  \sum_{n=1}^{N-1}\sum_{k=0}^{K-1}
    \Expect\left\|\widehat G_{\mathsf A}^{(n,k)}
      -\Expect[\widehat G_{\mathsf A}^{(n,k)}\mid\mathscr F_{n,k}]
                                       \right\|^2.
 \label{eq:batch-all-inner-transfer}
\end{align}
The coefficient $392$ equals $3\cdot16\cdot8+8$. Inserting~\eqref{eq:tic-batch-variance} proves~\eqref{eq:batch-tic-convergence}, and inserting~\eqref{eq:grpo2-batch-variance} proves~\eqref{eq:batch-grpo2-convergence}.

\subsection{Separate step-size optimization}

For a positive initial gap, each right-hand side in~\eqref{eq:batch-tic-convergence} and~\eqref{eq:batch-grpo2-convergence} is strictly convex in $\eta>0$, being a positive multiple of $1/\eta$ plus a positive linear term and a nonnegative quadratic term. It diverges at both ends of this interval. For TIC, its unique positive stationary point solves
\begin{align*}
 \frac{24[J^*-J(\theta_{1,0})]}{K(N-1)}
 &=\left(1-\frac1{|G|}\right)^2 R^2Lc_+
       \left(\frac1B+\frac{32\log|\mathcal V|}{|G|}\right)\\
 &\quad\times\left[48RLK\eta^2+784R^2L^2K(K-1)\eta^3\right].
\end{align*}
The GRPO$_2$ equation has one additional factor $T$ on the right. For each algorithm, the minimizer over the sufficient step interval is its own stationary point capped at $(128RLK)^{-1}$.

To obtain closed bounds with explicit parameter dependence, observe on this interval that
\[
 48RLK\eta+392R^2L^2K(K-1)\eta^2\leq52RLK\eta.
\]
For TIC choose the admissible witness
\[
 \eta=\frac1{RLK}\min\left\{\frac1{128},
 \sqrt{\frac{24[J^*-J(\theta_{1,0})]}
 {52R(N-1)(1-|G|^{-1})^2c_+
       (B^{-1}+32\log|\mathcal V|/|G|)}}\right\}.
\]
For GRPO$_2$, insert an additional factor $T$ in the denominator inside the square root. If the square-root choice is admissible, the inverse-step and linear terms are equal. If the cap is active, the linear term at the cap is at most the geometric mean of their two coefficients. These two cases show that the respective minimized bounds are at most
\begin{align*}
 &\frac{3072RLT[J^*-J(\theta_{1,0})]}{N-1}\\
 &\quad+72\left(1-\frac1{|G|}\right)R^{3/2}LT
  \sqrt{\frac{c_+[J^*-J(\theta_{1,0})]}{N-1}
                  \left(\frac1B+\frac{32\log|\mathcal V|}{|G|}\right)}
\end{align*}
for TIC, and the same expression with one additional $T$ inside the square root for GRPO$_2$. Here $2\sqrt{24\cdot52}<72$. Using $J^*-J(\theta_{1,0})\leq2R$ and $72\sqrt2<102$ proves~\eqref{eq:batch-tic-rate}--\eqref{eq:batch-grpo2-rate}. For a zero initial gap, replace the gap by $2R$ in the witness step choices. They are then strictly positive and give the same uniform bounds directly. All step choices refer to the respective algorithm's own run and deterministic bound.

\section{Per-response normalization and GRPO}
\label{su_grpo}

\begin{theorem}[GRPO convergence]\label{thm:grpo-stationarity}
Under Assumptions~\ref{L_smooth}--\ref{bounded_reward} and the conditional sampling protocol, let $N\geq2$, $B\geq1$, $|G|\geq2$, $\eta>0$, and $\epsilon_{\low},\epsilon_{\high}>0$. For $K\geq2$, original GRPO satisfies
\begin{align}
 &\frac1{(N-1)K}\sum_{n=1}^{N-1}\sum_{k=0}^{K-1}
       \Expect\|\nabla J(\theta_{n,k})\|^2\notag\\
 &\leq\frac{4T\bigl(J^*-J(\theta_{1,0})\bigr)}{\eta(N-1)K}
 +\frac{16R^3L^2T^2\eta}{\mathrm e(N-1)}
       \sum_{n=1}^{N-1}\Expect[\mathcal M_n^2]\notag\\
 &\quad+8TR^2(LT)^{3/2}\eta
       \left(3+\max\left\{1,\frac{3}{2\min\{\epsilon_{\low},\epsilon_{\high}\}}\right\}\right)
       \sqrt{\frac{(K-1)(2K-1)}6}\notag\\
 &\hspace{8em}\times
       \sqrt{\frac{4R^2L}{\mathrm e(N-1)}\sum_{n=1}^{N-1}\Expect[\mathcal M_n^2]}\notag\\
 &\quad+\frac{8R^2LT}{N-1}\left(\sum_{t=1}^{T}\frac1{t^2}\right)
       \sum_{n=1}^{N-1}\Expect\left[
          \frac{\sigma_{\theta_{n,0}}^2}{\Expect_{\theta_{n,0}}[|s_T|]}
       \right].\label{eq:conv_grpo}
\end{align}
The inverse-probability moments are evaluated along the run at step size $\eta$. The bound is interpreted in the extended real sense, allowing infinite inverse-probability moments.
For $K=1$, every update is on policy and
\begin{align}
 &\frac1{N-1}\sum_{n=1}^{N-1}\Expect\|\nabla J(\theta_{n,0})\|^2\notag\\
 &\leq\frac{4T[J^*-J(\theta_{1,0})]}{\eta(N-1)}
   +8R^3L^2T^2\eta\log|\mathcal V|\sum_{t=1}^{T}t^{-2}\notag\\
 &\quad+\frac{8R^2LT}{N-1}\left(\sum_{t=1}^{T}t^{-2}\right)
       \sum_{n=1}^{N-1}\Expect\!\left[
       \frac{\sigma_{\theta_{n,0}}^2}{\Expect_{\theta_{n,0}}[|s_T|]}\right].
 \label{eq:grpo-on-policy-explicit}
\end{align}
For a positive initial gap, the step size
\[
 \eta=\sqrt{\frac{J^*-J(\theta_{1,0})}
 {2R^3L^2T(N-1)\log|\mathcal V|\sum_{t=1}^{T}t^{-2}}}
\]
minimizes the first two terms of~\eqref{eq:grpo-on-policy-explicit}, giving
\begin{align}
 &\frac1{N-1}\sum_{n=1}^{N-1}\Expect\|\nabla J(\theta_{n,0})\|^2\notag\\
 &\leq8\sqrt2\,R^{3/2}LT^{3/2}
 \sqrt{\frac{[J^*-J(\theta_{1,0})]\log|\mathcal V|
                   \sum_{t=1}^{T}t^{-2}}{N-1}}\notag\\
 &\quad+\frac{8R^2LT}{N-1}\left(\sum_{t=1}^{T}t^{-2}\right)
       \sum_{n=1}^{N-1}\Expect\!\left[
       \frac{\sigma_{\theta_{n,0}}^2}{\Expect_{\theta_{n,0}}[|s_T|]}\right].
 \label{eq:grpo-on-policy-rate}
\end{align}
For zero initial gap, $\eta=1/\sqrt{T(N-1)}$ gives
\begin{align}
 &\frac1{N-1}\sum_{n=1}^{N-1}\Expect\|\nabla J(\theta_{n,0})\|^2\notag\\
 &\leq\frac{8R^3L^2T^{3/2}\log|\mathcal V|}{\sqrt{N-1}}
           \sum_{t=1}^{T}t^{-2}\notag\\
 &\quad+\frac{8R^2LT}{N-1}\left(\sum_{t=1}^{T}t^{-2}\right)
       \sum_{n=1}^{N-1}\Expect\!\left[
       \frac{\sigma_{\theta_{n,0}}^2}{\Expect_{\theta_{n,0}}[|s_T|]}\right].
 \label{eq:grpo-zero-gap-rate}
\end{align}
Each length term is evaluated along the run generated by its respective step choice. If $|\mathcal V|=1$, the objective gradient and initial gap are zero, so only the zero-gap branch applies.
\end{theorem}

Away from a clipping threshold, the multiplier in the first expression of Eq.~\eqref{eq:grpo_core} is
\[
 m_t^{\mathrm{GRPO}}=
 \begin{cases}
 \rho_t\,\mathbf1\{\rho_t\leq1+\epsilon_{\high}\},&A_c(s_T)\geq0,\\
 \rho_t\,\mathbf1\{\rho_t\geq1-\epsilon_{\low}\},&A_c(s_T)<0.
 \end{cases}
\]
At a threshold the multiplier is any value between the adjacent one-sided values. Thus $0\leq m_t^{\mathrm{GRPO}}\leq\rho_t$, including thresholds. The algorithm averages $Z_{\mathrm{GRPO}}$ over the actual prompt and response weights.

\begin{lemma}[The old-policy GRPO mean]\label{lem:grpo-global-mean}
For positive clipping margins,
\begin{align}
 &\Expect\left[\widehat G_{\mathrm{GRPO}}^{(k)}(\theta_{n,0},\theta_{n,0})
                 \middle|\mathscr F_{n,k}\right]\notag\\
 &=\left(1-\frac1{|G|}\right)
   \frac{\nabla J(\theta_{n,0})}{\Expect_{\theta_{n,0}}[|s_T|]}\notag\\
 &\quad+\left(1-\frac1{|G|}\right)
  \Expect_{\theta_{n,0}}\left[
    \left(\frac1{|s_T|}-\frac1{\Expect_{\theta_{n,0}}[|s_T|]}\right)
    (r(s_T)-\mu_{\theta_{n,0},c})G_T^{\theta_{n,0}}\right].\label{eq:grpo-global-mean}
\end{align}
The squared norm of the last expectation, including its prefactor, is at most
\begin{equation}
 4\left(1-\frac1{|G|}\right)^2 R^2L
   \left(\sum_{t=1}^{T}\frac1{t^2}\right)
   \frac{\sigma_{\theta_{n,0}}^2}
        {(\Expect_{\theta_{n,0}}[|s_T|])^2}.
 \label{eq:grpo-length-error}
\end{equation}
\end{lemma}
\begin{proof}
At the old parameter every ratio equals one and lies in the interior of each active clipping branch. Expand the empirical reward mean in each group's update. Among the $|G|^2$ pairs of a reward and a trajectory score divided by length, $|G|$ pairs use the same response and $|G|(|G|-1)$ use independent responses from the same prompt. This gives exactly
\[
 \left(1-\frac1{|G|}\right)
 \Expect_{\theta_{n,0}}\left[
       (r(s_T)-\mu_{\theta_{n,0},c})\frac{G_T^{\theta_{n,0}}}{|s_T|}
 \right].
\]
Adding and subtracting the reciprocal joint mean length gives~\eqref{eq:grpo-global-mean}, because the old trajectory score has conditional mean zero given its prompt. For the remainder, write the reciprocal difference as
\[
 \frac{\Expect_{\theta_{n,0}}[|s_T|]-|s_T|}
      {|s_T|\Expect_{\theta_{n,0}}[|s_T|]}.
\]
In any direction, Cauchy--Schwarz, $|r-\mu_{\theta_{n,0},c}|\leq2R$, and~\eqref{eq:stopped-second} give~\eqref{eq:grpo-length-error}. Both the expectation and the variance are taken jointly over prompts and responses.
\end{proof}

\begin{lemma}[GRPO drift and second moment]\label{lem:grpo-drift}
For the same empirical group and positive clipping margins,
\begin{align}
 &\left\|\Expect\left[
   \widehat G_{\mathrm{GRPO}}^{(k)}(\theta_{n,k},\theta_{n,0})
  -\widehat G_{\mathrm{GRPO}}^{(k)}(\theta_{n,0},\theta_{n,0})
   \middle|\mathscr F_{n,k}\right]\right\|\notag\\
 &\leq2RLT\left(2+\max\left\{1,
          \frac3{2\min\{\epsilon_{\low},\epsilon_{\high}\}}\right\}\right)
       \|\theta_{n,k}-\theta_{n,0}\|,\label{eq:grpo-drift}\\
 &\left\|\widehat G_{\mathrm{GRPO}}^{(n,k)}\right\|^2
       \leq\frac{4R^2L}{\mathrm e}\mathcal M_n^2.
       \label{eq:grpo-second}
\end{align}
\end{lemma}
\begin{proof}
At a fixed active prefix, integrate the ratio-score product along the segment from $\theta_{n,0}$ to $\theta_{n,k}$. Its derivative is the sum of the ratio times the score derivative and the ratio times the outer product of the score applied to the parameter displacement. In the conditional old-policy expectation of the absolute projection onto any direction, change the next-action law using the ratio. The Lipschitz score bound controls the first term by $L\|\theta_{n,k}-\theta_{n,0}\|$, and the directional Fisher bound controls the second by the same quantity. Hence
\[
 \sup_{\|\theta'-\theta\|=1}
 \Expect_{\theta_{n,0}}\left[
  \left|\left\langle\theta'-\theta,
       \rho_tg_t^{\theta_{n,k}}-g_t^{\theta_{n,0}}\right\rangle\right|
     \middle|s_{t-1}\right]
 \leq2L\|\theta_{n,k}-\theta_{n,0}\|.
\]
On an inactive branch, including a threshold with a fractional active coefficient, the ratio differs from one by at least
$\min\{\epsilon_{\low},\epsilon_{\high}\}$. Express the ratio difference as the integral of its derivative along the same segment. Markov's inequality applied pointwise to this absolute integral, followed by a next-action change of measure, Cauchy--Schwarz, and score Lipschitzness, bounds the expected absolute old-score projection on that event by
\[
 \frac{L\|\theta_{n,k}-\theta_{n,0}\|}
      {\min\{\epsilon_{\low},\epsilon_{\high}\}}
 \left(1+\frac{\sqrt L\|\theta_{n,k}-\theta_{n,0}\|}{2}\right).
\]
When $\sqrt L\|\theta_{n,k}-\theta_{n,0}\|\leq1$, this is at most the factor $3/(2\min\{\epsilon_{\low},\epsilon_{\high}\})$ times $L\|\theta_{n,k}-\theta_{n,0}\|$. Otherwise the unconditional old-score first-moment bound $\sqrt L$ is at most $L\|\theta_{n,k}-\theta_{n,0}\|$. These two bounds control the inactive part globally. The active part is bounded by the preceding ratio-score estimate. Finally use $|A_c|\leq2R$, $|s_T|^{-1}\leq1$, and at most $T$ active tokens, before taking expectations. The branch-event inclusion and reward bounds hold pathwise, so the conditional estimates permit dependence of the active branch on the same group's empirical advantage. They prove~\eqref{eq:grpo-drift}.

For~\eqref{eq:grpo-second}, self-bounding gives
\[
 \|\nabla\Pro_\theta(s_t\mid s_{t-1})\|^2
 \leq-2L\Pro_\theta(s_t\mid s_{t-1})^2
                 \log\Pro_\theta(s_t\mid s_{t-1})\leq L/\mathrm e.
\]
The last inequality follows by maximizing the scalar function
$-x^2\log x$ on $(0,1]$, whose maximum is $1/(2\mathrm e)$.
Use $0\leq m_t^{\mathrm{GRPO}}\leq\rho_t$ and Jensen's inequality for the per-response token average and the empirical response and prompt averages. This yields the stated pathwise bound with the maximum over the sampled transitions. The resulting moment bound retains this random maximum inside the expectation.
\end{proof}

\subsection{Proof of Theorem~\ref{thm:grpo-stationarity}}
\begin{proof}
First let $K\geq2$. If the displayed inverse-probability moment is infinite,~\eqref{eq:conv_grpo} holds in the extended real sense. Otherwise, the moment is finite and the following estimates apply. Pair the old-policy identity~\eqref{eq:grpo-global-mean} with the current gradient. Young's inequality assigns half of the coefficient
$(1-|G|^{-1})/\Expect_{\theta_{n,0}}|s_T|$ to the length remainder. Use~\eqref{eq:grpo-length-error},~\eqref{eq:grpo-drift}, and the second bound in Eq.~\eqref{eq:gradient-envelope}. Since $|G|\geq2$ and the mean length is at most $T$, this gives
\begin{align*}
 &\left\langle\nabla J(\theta_{n,k}),
      \Expect[\widehat G_{\mathrm{GRPO}}^{(n,k)}\mid\mathscr F_{n,k}]\right\rangle
 \geq\frac{\|\nabla J(\theta_{n,k})\|^2}{4T}\\
 &\quad-2R^2(LT)^{3/2}
      \left(3+\max\left\{1,\frac3{2\min\{\epsilon_{\low},\epsilon_{\high}\}}\right\}\right)
      \|\theta_{n,k}-\theta_{n,0}\|\\
 &\quad-2R^2L\left(\sum_{t=1}^{T}\frac1{t^2}\right)
           \frac{\sigma_{\theta_{n,0}}^2}{\Expect_{\theta_{n,0}}[|s_T|]}.
\end{align*}
Here the first bound in Eq.~\eqref{eq:gradient-envelope} controls the current-gradient norm when pairing it with the drift. The objective ascent inequality adds a cost at most
$RLT\eta^2\Expect[\|\widehat G_{\mathrm{GRPO}}^{(n,k)}\|^2\mid\mathscr F_{n,k}]$.

Summing the inner-update displacements and applying Cauchy--Schwarz gives
\begin{align*}
 &\frac1{(N-1)K}\sum_{n=1}^{N-1}\sum_{k=0}^{K-1}
       \Expect\|\theta_{n,k}-\theta_{n,0}\|\\
 &\leq\eta\sqrt{\frac{(K-1)(2K-1)}6}
       \sqrt{\frac1{(N-1)K}\sum_{n=1}^{N-1}\sum_{k=0}^{K-1}
                     \Expect\|\widehat G_{\mathrm{GRPO}}^{(n,k)}\|^2}\\
 &\leq\eta\sqrt{\frac{(K-1)(2K-1)}6}
       \sqrt{\frac{4R^2L}{\mathrm e(N-1)}
                        \sum_{n=1}^{N-1}\Expect\mathcal M_n^2}.
\end{align*}
For the first inequality, each update with index $j$ appears in the later displacements $K-1-j$ times; the average of the squares of these coefficients is $(K-1)(2K-1)/6$. Sum the ascent inequalities over all $(N-1)K$ updates, telescope $J$, bound the total increase by $J^*-J(\theta_{1,0})$, and multiply by $4T/\eta$. Substitution of~\eqref{eq:grpo-second} proves~\eqref{eq:conv_grpo}.

When $K=1$, every update is on policy. Empirical covariance Cauchy--Schwarz gives, pathwise,
\[
 \left\|\widehat G_{\mathrm{GRPO}}^{(n,0)}\right\|^2
 \leq R^2\frac1B\sum_{\text{prompt occurrences}}
       \sum_{s_T\in\mathcal S_T}\xi_c(s_T)
          \frac{\|G_T^{\theta_{n,0}}\|^2}{|s_T|^2}.
\]
The trace form of~\eqref{eq:stopped-second} bounds its expectation by
$2R^2L\log|\mathcal V|\sum_{t=1}^{T}t^{-2}$. Substitution into the same ascent argument proves~\eqref{eq:grpo-on-policy-explicit}.

The sum $\sum_{t=1}^{T}t^{-2}$ is at most $2$, by comparison of its
terms after the first with the integral of $x^{-2}$ over $[1,\infty)$.
Absorbing only $R,L,K$, the clipping margins, and the initial gap into
constants in~\eqref{eq:conv_grpo} proves~\eqref{eq:grpo-order}.
For $K=1$, its step-dependent part is
\[
 \frac{4T[J^*-J(\theta_{1,0})]}{\eta(N-1)}
 +8R^3L^2T^2\eta\log|\mathcal V|\sum_{t=1}^{T}\frac1{t^2}.
\]
For positive initial gap, differentiation or the arithmetic--geometric
mean inequality gives the step size in
Theorem~\ref{thm:grpo-stationarity}. Its value is
\[
 8\sqrt2\,R^{3/2}LT^{3/2}
 \sqrt{\frac{[J^*-J(\theta_{1,0})]\log|\mathcal V|
                   \sum_{t=1}^{T}t^{-2}}{N-1}}.
\]
Adding the length term proves~\eqref{eq:grpo-on-policy-rate}; that term remains evaluated along the run at the selected step size.
For zero initial gap, substituting $\eta=1/\sqrt{T(N-1)}$ into~\eqref{eq:grpo-on-policy-explicit} proves~\eqref{eq:grpo-zero-gap-rate}. Both step choices imply~\eqref{eq:grpo-on-policy-order}.

Finally, averaging $(|s_T|-1)(T-|s_T|)\geq0$ under the joint old law gives
\begin{align*}
 \frac{\sigma_{\theta_{n,0}}^2}{\Expect_{\theta_{n,0}}[|s_T|]}
 &\leq T+1-\Expect_{\theta_{n,0}}[|s_T|]
                 -\frac{T}{\Expect_{\theta_{n,0}}[|s_T|]}\\
 &\leq(\sqrt T-1)^2.
\end{align*}
The second inequality is the arithmetic--geometric mean inequality.
Thus the individual-length contribution in~\eqref{eq:conv_grpo} is at
most $16R^2LT(\sqrt T-1)^2$, as stated in the main text.
\end{proof}

\section{Conditional core moments}
\label{app:proof_variance_upper_bounds}

\begin{proposition}[Core update moments]\label{prop:variance_upper_bounds}
Under the stated sampling protocol and Assumptions~\ref{L_smooth}--\ref{bounded_reward},
\begin{equation}
\label{eq:grpo_core_upper}
\begin{aligned}
&\Expect[(\|Z_{\mathrm{GRPO}}\|^2,\|Z_{\mathrm{GRPO}_2}\|^2,\|Z_{\mathrm{TIC}}\|^2)\mid\mathscr F_{n,k}]\\
&\qquad\leq4R^2L\left(\frac{\Expect[\mathcal M_n^2\mid\mathscr F_{n,k}]}{\mathrm e},2c_+\log|\mathcal V|,\frac{2c_+\log|\mathcal V|}{T}\right).
\end{aligned}
\end{equation}
Expectation and inequality are interpreted componentwise, in the extended real sense for the first component.
Here the distinguished response is sampled using the empirical prompt and response weights. The same upper bounds apply to the empirical average update by Jensen's inequality.
\end{proposition}

\begin{proof}[Proof of Proposition~\ref{prop:variance_upper_bounds}]
For GRPO, $|A_c|\leq2R$, the inequality $0\leq m_t^{\mathrm{GRPO}}\leq\rho_t$, and the probability-gradient bound in Lemma~\ref{lem:grpo-drift} give the pathwise estimate
\[
 \|Z_{\mathrm{GRPO}}\|^2\leq
 \frac{4R^2}{|s_T|}\sum_{t=1}^{|s_T|}
          \frac{L}{\mathrm e\Pro_{\theta_{n,0}}(s_t\mid s_{t-1})^2}
 \leq\frac{4R^2L}{\mathrm e}\mathcal M_n^2.
\]
Take conditional expectations to obtain the first componentwise inequality in Eq.~\eqref{eq:grpo_core_upper}.

For GRPO$_2$, the bound $|A_c|\leq2R$ and~\eqref{eq:grpo2-feature-moments} give
\[
 \Expect[\|Z_{\mathrm{GRPO}_2}\|^2\mid\mathscr F_{n,k}]
 \leq\frac{4R^2}{T^2}\,2c_+LT^2\log|\mathcal V|
 =8R^2c_+L\log|\mathcal V|.
\]
This is the second componentwise inequality in Eq.~\eqref{eq:grpo_core_upper}.
For TIC,~\eqref{eq:tic-feature-moments} similarly gives
\[
 \Expect[\|Z_{\mathrm{TIC}}\|^2\mid\mathscr F_{n,k}]
 \leq\frac{4R^2}{T^2}\,2c_+LT\log|\mathcal V|
 =\frac{8R^2c_+L\log|\mathcal V|}{T}.
\]
This is the third componentwise inequality in Eq.~\eqref{eq:grpo_core_upper}.
The empirical advantage is bounded pathwise before taking expectations, so it need not be independent of the score. Jensen's inequality for the nonnegative empirical weights extends the bounds to the averaged updates.
\end{proof}

\section{Hard-instance proofs for the variance bounds}
\label{app:hard_instance_variance}

\begin{proposition}[Response-level variance lower bounds]
\label{prop:hard_instance_variance}
Fix $R,L>0$ and $c_+\geq1$. For every $T\geq2$ and $|G|\geq2$, there are separate one-dimensional binary policy instances satisfying Assumptions~\ref{L_smooth}--\ref{bounded_reward}, with one prompt, independent responses, and deterministic response length $T$, for which
\begin{align}
 \Expect\!\left[\left\|Z_{\mathrm{GRPO}_2}-\Expect[Z_{\mathrm{GRPO}_2}\mid\mathscr F_{n,k}]\right\|^2\mid\mathscr F_{n,k}\right]
 &\geq\frac{R^2L(|G|-1)}{256|G|},\label{eq:grpo2_core_lower}\\
 \Expect\!\left[\left\|Z_{\mathrm{TIC}}-\Expect[Z_{\mathrm{TIC}}\mid\mathscr F_{n,k}]\right\|^2\mid\mathscr F_{n,k}\right]
 &\geq\frac{R^2L(|G|-1)}{8|G|T}.\label{eq:tic_core_lower}
\end{align}
For original GRPO, a binary instance with $|G|=2$ satisfies, whenever $T\geq64$ and $T^2>2c_+$,
\begin{equation}
 \Expect\!\left[\left\|Z_{\mathrm{GRPO}}-\Expect[Z_{\mathrm{GRPO}}\mid\mathscr F_{n,k}]\right\|^2\mid\mathscr F_{n,k}\right]
 \geq\frac{R^2LT}{256}.
 \label{eq:grpo_core_lower}
\end{equation}
Here the cores are evaluated pointwise at fixed old/current parameter pairs before sampling. In the original-GRPO construction, the old-policy success probability is $T^{-2}$; in the GRPO$_2$ construction it is $1/(4c_+)$, independently of $T$.
\end{proposition}

\begin{proof}[Proof of Proposition~\ref{prop:hard_instance_variance}]
Use one prompt and a scalar parameter. At every parameter-dependent transition, let the probability of token $1$ be
\[
 \frac{\exp(2\sqrt L\,\theta)}{1+\exp(2\sqrt L\,\theta)}.
\]
Its score is $2\sqrt L$ times the difference between the observed binary token and its success probability. The derivative of this score has absolute value
\[
 4L\frac{\exp(2\sqrt L\,\theta)}{(1+\exp(2\sqrt L\,\theta))^2}\leq L.
\]
Thus the score is globally $L$-Lipschitz, and both token probabilities are positive for every finite parameter. Responses have exactly $T$ tokens. All groups consist of independent responses, and the parameter pairs below are fixed before sampling.

\paragraph{GRPO$_2$.}
Let the first $T-1$ tokens follow this policy independently. The last token is an independent, parameter-independent fair binary token, and the terminal reward is $R$ times that token. This reward lies in $[0,R]$ and the last score is zero. Set
\[
 \theta_{n,k}=0,\qquad
 \theta_{n,0}=-\frac{\log(4c_+-1)}{2\sqrt L}.
\]
The current success probability is $1/2$, while the old success probability is $1/(4c_+)$. For $t\leq T-1$, the score at the current parameter is $\sqrt L$ on token $1$ and $-\sqrt L$ on token $0$. The clipped ratio is $c_+$ on token $1$ and $2c_+/(4c_+-1)$ on token $0$. Hence, directly under the old response law,
\[
 \Expect[\rho_t^+g_t^{\theta_{n,k}}\mid\mathscr F_{n,k}]
 =\frac{c_+\sqrt L}{4c_+}
  -\left(1-\frac1{4c_+}\right)\frac{2c_+\sqrt L}{4c_+-1}
 =-\frac{\sqrt L}{4}.
\]
The empirical advantage of a distinguished response depends only on the group's last tokens. It is independent of that response's first $T-1$ tokens and has moments
\begin{align*}
 \Expect[A_c(s_T)\mid\mathscr F_{n,k}]&=0,\\
 \Expect[A_c(s_T)^2\mid\mathscr F_{n,k}]
 &=\frac{R^2}{4}\left[\left(1-\frac1{|G|}\right)^2+
                \frac{|G|-1}{|G|^2}\right]
 =\frac{R^2(|G|-1)}{4|G|}.
\end{align*}
Consequently $\Expect[Z_{\mathrm{GRPO}_2}\mid\mathscr F_{n,k}]=0$. Jensen's inequality applied to the full token sum gives
\begin{align*}
 \Expect[\|Z_{\mathrm{GRPO}_2}\|^2\mid\mathscr F_{n,k}]
 &=\frac{R^2(|G|-1)}{4|G|T^2}
   \Expect\!\left[\left(\sum_{t=1}^{T-1}\rho_t^+g_t^{\theta_{n,k}}\right)^2\mid\mathscr F_{n,k}\right]\\
 &\geq\frac{R^2L(|G|-1)(T-1)^2}{64|G|T^2}
 \geq\frac{R^2L(|G|-1)}{256|G|}.
\end{align*}
This proves~\eqref{eq:grpo2_core_lower} for every $T\geq2$.

\paragraph{TIC-GRPO.}
Keep the same independent reward token and empirical baseline, and set $\theta_{n,k}=\theta_{n,0}=0$. All likelihood ratios equal one. The first $T-1$ scores are independent, have mean zero, and have second moment $L$. The empirical advantage remains independent of their sum. Therefore the core has mean zero and
\[
 \Expect[\|Z_{\mathrm{TIC}}\|^2\mid\mathscr F_{n,k}]
 =\frac{R^2L(|G|-1)(T-1)}{4|G|T^2}
 \geq\frac{R^2L(|G|-1)}{8|G|T},
\]
which is~\eqref{eq:tic_core_lower}.

\paragraph{Original GRPO.}
Let all $T$ tokens follow the parameter-dependent policy independently and let $r(s_T)=R$ for the all-zero response and zero otherwise. Use $|G|=2$ and set
\[
 \theta_{n,k}=0,\qquad
 \theta_{n,0}=-\frac{\log(T^2-1)}{2\sqrt L}.
\]
The old success probability is $T^{-2}$. Consider the event that the distinguished response has exactly one token $1$ and the other response is all zeros. Its probability is
\[
 \frac1T(1-T^{-2})^{2T-1}
 \geq\frac1T\left(1-\frac{2T-1}{T^2}\right)
 \geq\frac1{2T}\qquad(T\geq4).
\]
On this event the distinguished advantage equals $-R/2$. Its token $1$ has ratio $T^2/2>c_+$ and score $\sqrt L$, so the negative-advantage branch retains that ratio. Each zero token has score $-\sqrt L$ and multiplier between zero and $1/[2(1-T^{-2})]$. Thus
\[
 \sum_{t=1}^T m_t^{\mathrm{GRPO}}g_t^{\theta_{n,k}}
 \geq\sqrt L\left(\frac{T^2}{2}-\frac{T-1}{2(1-T^{-2})}\right)
 =\frac{\sqrt L\,T^3}{2(T+1)}
 \geq\frac{\sqrt L\,T^2}{4}.
\]
It follows that the squared core norm on this event is at least $R^2LT^2/64$, and hence
\[
 \Expect[\|Z_{\mathrm{GRPO}}\|^2\mid\mathscr F_{n,k}]
 \geq\frac{R^2LT}{128}.
\]
To bound the mean, use $|A_c|\leq R/2$ for this two-response group, $0\leq m_t^{\mathrm{GRPO}}\leq\rho_t$, and $|g_t^{\theta_{n,k}}|=\sqrt L$. Each old-policy token ratio has expectation one, so
\[
 \left|\Expect[Z_{\mathrm{GRPO}}\mid\mathscr F_{n,k}]\right|
 \leq\frac{R}{2T}\sum_{t=1}^T
        \Expect[\rho_t|g_t^{\theta_{n,k}}|\mid\mathscr F_{n,k}]
 =\frac{R\sqrt L}{2}.
\]
Subtracting the squared mean yields
\begin{align*}
 &\Expect\!\left[\left\|Z_{\mathrm{GRPO}}-\Expect[Z_{\mathrm{GRPO}}\mid\mathscr F_{n,k}]\right\|^2\mid\mathscr F_{n,k}\right]\\
 &\qquad\geq R^2L\left(\frac{T}{128}-\frac14\right)
 \geq\frac{R^2LT}{256}\qquad(T\geq64).
\end{align*}
This proves~\eqref{eq:grpo_core_lower}.
\end{proof}

\section*{AI use statement}\label{sec:ai-use}
All mathematical proofs and derivations were independently developed and verified by the authors. AI assistance with mathematical verification was limited to generating Lean code and running Lean checks for selected components of the analysis. Generative AI tools (OpenAI ChatGPT and Codex) also assisted with literature discovery and reference verification, manuscript drafting, and \LaTeX{} preparation. The authors take full responsibility for the correctness, originality, and integrity of the paper, including all AI-assisted content.

\section*{Ethics statement}\label{sec:ethics}
This work studies the optimization of reward-based language-model training and evaluates public models on mathematical reasoning and coding benchmarks. The mathematical guarantees concern optimization and update variance; their scope does not include the safety or fairness of deployed language models.

\section*{Reproducibility statement}\label{sec:reproducibility}
Section~\ref{sec:variance_analysis} specifies the regularity assumptions and conditional sampling protocol. The update rules appear in Sections~\ref{sec:grpo_review} and~\ref{sec:tic_grpo}, and the appendices provide complete mathematical proofs and variance constructions. Appendix~\ref{sec:experiments} describes the models, datasets, training configuration, evaluation metrics, and component ablations, including the normalization and sampling conventions used in the experiments.

\clearpage
\endgroup

\end{document}